\documentclass{ieeetj}
\usepackage{cite}
\usepackage{amsmath,amssymb,amsfonts}
\usepackage{algorithmic}
\usepackage{graphicx,color}
\usepackage{textcomp}
\usepackage{xcolor}
\usepackage{hyperref}
\usepackage{svg}
\usepackage{algorithm,algorithmic}
\def\BibTeX{{\rm B\kern-.05em{\sc i\kern-.025em b}\kern-.08em
T\kern-.1667em\lower.7ex\hbox{E}\kern-.125emX}}
\AtBeginDocument{\definecolor{tmlcncolor}{cmyk}{0.93,0.59,0.15,0.02}\definecolor{NavyBlue}{RGB}{0,86,125}}

\def\OJlogo{\vspace{-4pt}$<$Society logo(s) and publication title will appear here.$>$}
\def\seclogo{\vspace{10pt}$<$Society logo(s) and publication title will appear here.$>$}

\usepackage[english]{babel}
\usepackage{graphicx}
\usepackage{multirow}
\usepackage{amsfonts}
\usepackage{tabularx}
\usepackage{algorithm,algorithmic}
\usepackage{amsthm}
\usepackage{bm}
\usepackage{enumerate}
\usepackage{adjustbox}
\usepackage{xcolor}
\usepackage{siunitx}
\usepackage{fix-cm}
\usepackage{mathtools}
\usepackage{array}
\usepackage{cite}
\makeatletter
\let\NAT@parse\undefined
\makeatother
\usepackage{wasysym}
\usepackage{url}
\usepackage{xurl}
\usepackage{booktabs}
\usepackage{hyperref}
\usepackage{cleveref}

\def\OJlogo{\vspace{-4pt}$<$Society logo(s) and publication title will appear here.$>$}
\def\seclogo{\vspace{10pt}$<$Society logo(s) and publication title will appear here.$>$}

\title{Time-Optimized Safe Navigation in Unstructured Environments through Learning Based Depth Completion}
\author{Jeffrey Mao$^1$, Raghuram Cauligi Srinivas$^1$,  Stephen Nogar$^2$,  and Giuseppe Loianno$^3$}
\affil{New York University, New York City, NY 11217 USA}
\affil{US DEVCOM Army Research Lab}
\affil{University of California, Berkeley, CA 94720, USA. }
\corresp{Corresponding author: Jeffrey Mao (email: jm7752@nyu.edu).}

\begin{document}
\receiveddate{May 2025}
\reviseddate{October 2025}
\accepteddate{XX Month, XXXX}
\publisheddate{XX Month, XXXX}
\currentdate{XX Month, XXXX}

\begin{abstract}
 Quadrotors hold significant promise for several applications such as agriculture, search-and-rescue, and infrastructure inspection. Achieving autonomous operation requires systems to navigate safely through complex and unfamiliar environments. This level of autonomy is particularly challenging due to the complexity of such environments and the need for real-time decision-making—especially for platforms constrained by size, weight, and power (SWaP), which limits flight time and precludes the use of bulky sensors like Light Detection and Ranging (LiDAR) for mapping. Furthermore, computing globally optimal, collision-free paths and translating them into time-optimized, safe trajectories in real time adds significant computational complexity.
To address these challenges, we present a fully onboard, real-time navigation system that relies solely on lightweight onboard sensors.
Our system constructs a dense 3D map of the environment using a novel visual depth estimation approach that fuses stereo and monocular learning-based depth, yielding longer-range, denser, and less noisy depth maps than conventional stereo methods.
Building on this map, we introduce a novel planning and trajectory generation framework capable of rapidly computing time-optimal global trajectories. 
As the map is incrementally updated with new depth information, our system continuously refines the trajectory to maintain safety and optimality. Both our planner and trajectory generator outperforms state-of-the-art methods in terms of computational efficiency and guarantee obstacle-free trajectories.
We validate our system through robust autonomous flight experiments in diverse indoor and outdoor environments, demonstrating its effectiveness for safe navigation in previously unknown settings.\end{abstract}

\maketitle

\section{Introduction} \label{sec:introduction}

\begin{figure}
    \centering
    \includegraphics[width=0.95\linewidth]{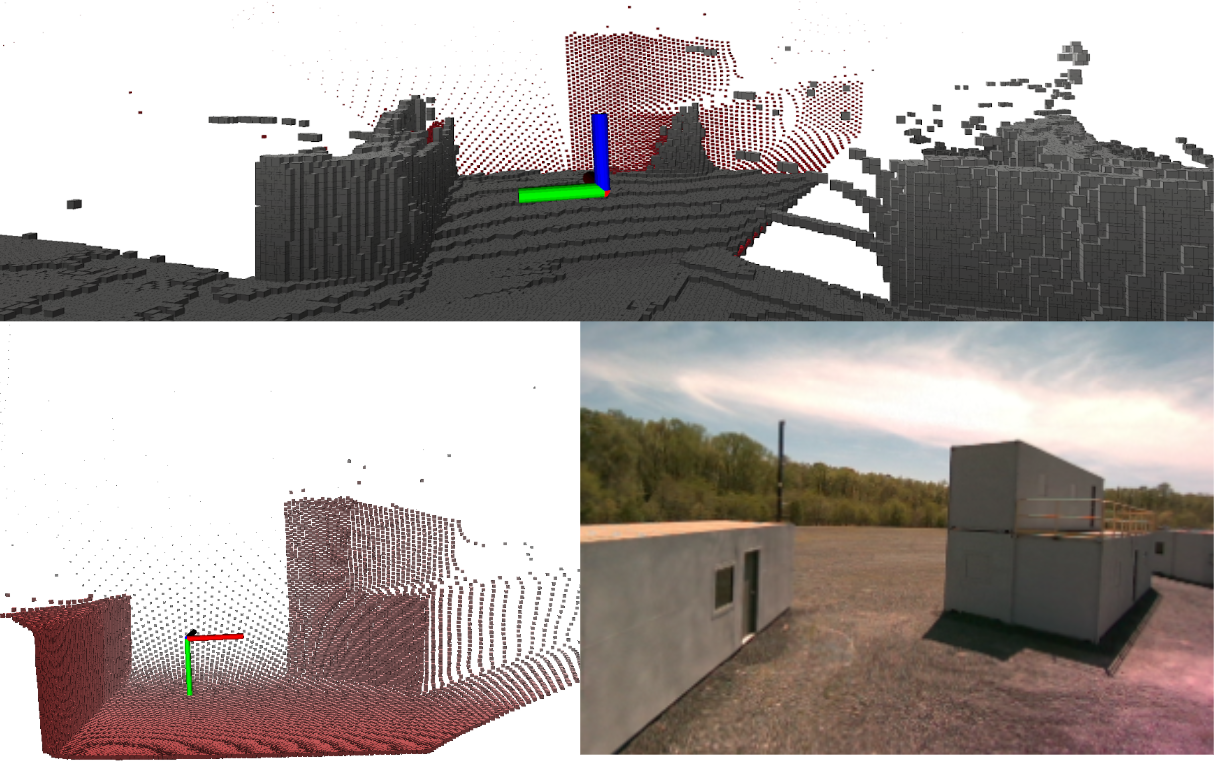}
    \vspace{-1em}
    \caption{A map of our environment on the top. Bottom right is the color image, and bottom Left is our depth completion.}
    \label{fig:intro_figure}
\end{figure}

\begin{table*}[t!]
    \centering
    \begin{tabular}{c|c|c|c|c|c|c|c|c}
      Method  & Devo\cite{alessandro_rl_explore} & MINCO \cite{zju_swarm}   & Saviolo \cite{saviolo2024reactivecollisionavoidancesafe} & \textbf{OUR} &   Rappids \cite{rappids}& Lee\cite{RapidQuad}  &  Super \cite{ren2025safety}   & Zhou \cite{zhou2019robust}   \\
      \hline
      Sensors & 1x RGB  & 1x RGBD  & 1x RGBD  &\textbf{1x RGBD} & 2x Stereo Pairs & 2x RGBD  & 1x LiDAR  & 1x LiDAR   \\
      Total Sensor Mass & 5 g & 15 g & 72 g & \textbf{100 g} & 127 g & 200 g    &  265 g & 840 g \\
     Sensor Range& N/A  & 3 m  &  8 m   & \textbf{10 m}& 3 m  & 6 m & 40 m  & 100 m\\
      Safety Guarantee & No &  No  & Yes & \textbf{Yes} & Yes & Yes & Yes & Yes\\
      Traj. Planning & Learning &  Global  & Local & \textbf{Global} & Local & Local & Global & Local \\
    \end{tabular}
    \caption{Autonomous Navigation Systems Comparisons. Safety Guarantees in our use case is defined as the trajectory solved can never intersect with an obstacle or closer than some safety distance. It should be noted that Super \cite{ren2025safety} trajectory generation is based on MINCO.}
    \label{tab:related_works}
    \vspace{-1em}
\end{table*}

\IEEEPARstart{A}{utonomous} Micro Aerial Vehicles (MAVs) offer valuable solutions across a range of application scenarios such as agriculture \cite{KRESTENITIS2024104581}, search-and-rescue \cite{search_rescue}, and infrastructure inspection \cite{inspection}.  
To services these applications, the aerial vehicles have to locate the obstacles in real time while concurrently generating and executing safe and obstacle free trajectories.

Many existing systems rely on heavyweight sensors such as Light Detection and Ranging (LiDAR) to ensure collision-free navigation~\cite{ren2025safety}. However, these sensors are often incompatible with the Size, Weight, and Power (SWaP) constraints of small-scale aerial vehicles, limiting flight time and agility. While lightweight alternatives like cameras are more suitable for SWaP-constrained platforms, they frequently struggle with noisy and inaccurate obstacle localization. Moreover, computing globally optimal, obstacle-free paths in large-scale environments is computationally intensive, and transforming these paths into time-optimal, safe trajectories in real time introduces additional complexity.

To address these challenges, we propose an autonomous quadrotor navigation framework that enables real-time exploration and mapping of unknown indoor and outdoor environments. The system operates using a lightweight sensor suite consisting of cameras, an IMU, and GPS (when available), making it well-suited for SWaP-constrained platforms.
Our navigation stack can generate high quality depth, convert it to an accurate map, and autonomously plan safe trajectories around obstacles illustrated in Fig.~\ref{fig:intro_figure}.
In this context, safety refers to the system's ability to provide guarantees for the planned trajectories by ensuring they avoid collisions with obstacles and stay within operational limits, based on the robot's current environment map.

\textbf{Contributions.} The  contributions of this work are
\begin{itemize}
    \item We develop a visual perception algorithm capable of generating dense and low-noise depth based on depth completion from learning based depth and classical computer vision.
    \item We propose the first real-time implementation for shortest path on the graph of convex sets (our path planning method) in unknown environments.
    \item We develop a unique time-optimized trajectory generation method with safety guarantees along with a mathematical derivation of our polynomial properties. Our method generates global trajectories more quickly while ensuring safety. 
    \item We integrate the proposed algorithms into a  complete software stack for real time onboard aerial autonomous navigation in unstructured environments that relies solely on lightweight onboard sensors, namely cameras and an IMU.  
    \item We extensively validate our navigation system in both indoor (GPS denied) and outdoor environments for exploration tasks under varying lighting conditions and harsh winds of up to $15\si{mph}$, demonstrating reliable performance over areas of up to $60\si{m}\times40\si{m}\times3\si{m}$.
\end{itemize}


\section{Related Works} \label{sec:related_works}
Autonomously Navigating environment requires two major components. First the vehicle needs to be able to localize the obstacles in the environments. Next, the vehicle needs to intelligently plan a path that avoids obstacles. Several related works featuring both components are listed in Table~\ref{tab:related_works}. Overall, our approach offers the best sensing range among vision-based methods using lightweight cameras, while maintaining safety guarantees for the planned trajectories. In the following, we first focus on describing relevant works in the mapping area and then planning approaches.

\textbf{Mapping}. There are two predominant sensors that are currently employed to generate environment depth onboard a quadrotor: cameras or LiDAR.  Although other sensors exist, such as radar\cite{hong2021radar} or ultrasounds \cite{ultrasound}, they are far less prevalent on drones due to their size or low resolution. 
Most works using LiDARs \cite{zhang2014loam,zhou2019robust, ren2025safety,  zheng2022fast} are able to achieve strong mapping and localization performance. However, the LiDAR's weight ranges from $250~\si{g}$ to $1000~\si{g}$ which limits their use on smaller aerial platforms. 
In contrast, stereo cameras offer a much lighter payload, typically less than $100~\si{g}$. However small stereo camera pairs have low maximum depth sensing range, typically $3~\si{m}-6~\si{m}$ \cite{zju_swarm,rappids,RapidQuad}. 
Learning based monocular depth estimation methods~\cite{depthanything} offer an exciting new paradigm. Monocular depth estimation inferences depth based on the image's structure by learning the mapping from images to depth on large datasets.
Unfortunately, while the depth structure is relatively consistent, the scale of the depth is wildly off for objects outside the training set. 
One popular paradigm for addressing this problem is called depth completion, where a sparse metric depth supplements monocular information to create a dense structurally consistent metric depth. 
A supplemental sensor like a radar \cite{li2024radarcam} can be fed into a larger neural network to infer the scale. However, this solution involves large datasets of unified sensor data to create this depth completion method therefore limiting their deployment.
Unlike those methods, in the proposed work we develop an optimization-based scaling method using our sparse depth generated from our RGB-D camera similar to \cite{saviolo2024reactivecollisionavoidancesafe}, but we employ inverse depth obtaining more accurate and consistent results. 
This method does not require retraining and is applicable to a variety of settings giving us great flexibility for deployment and switching the components in this system. 

\begin{figure*}[t]
\centering\includegraphics[width=\linewidth]{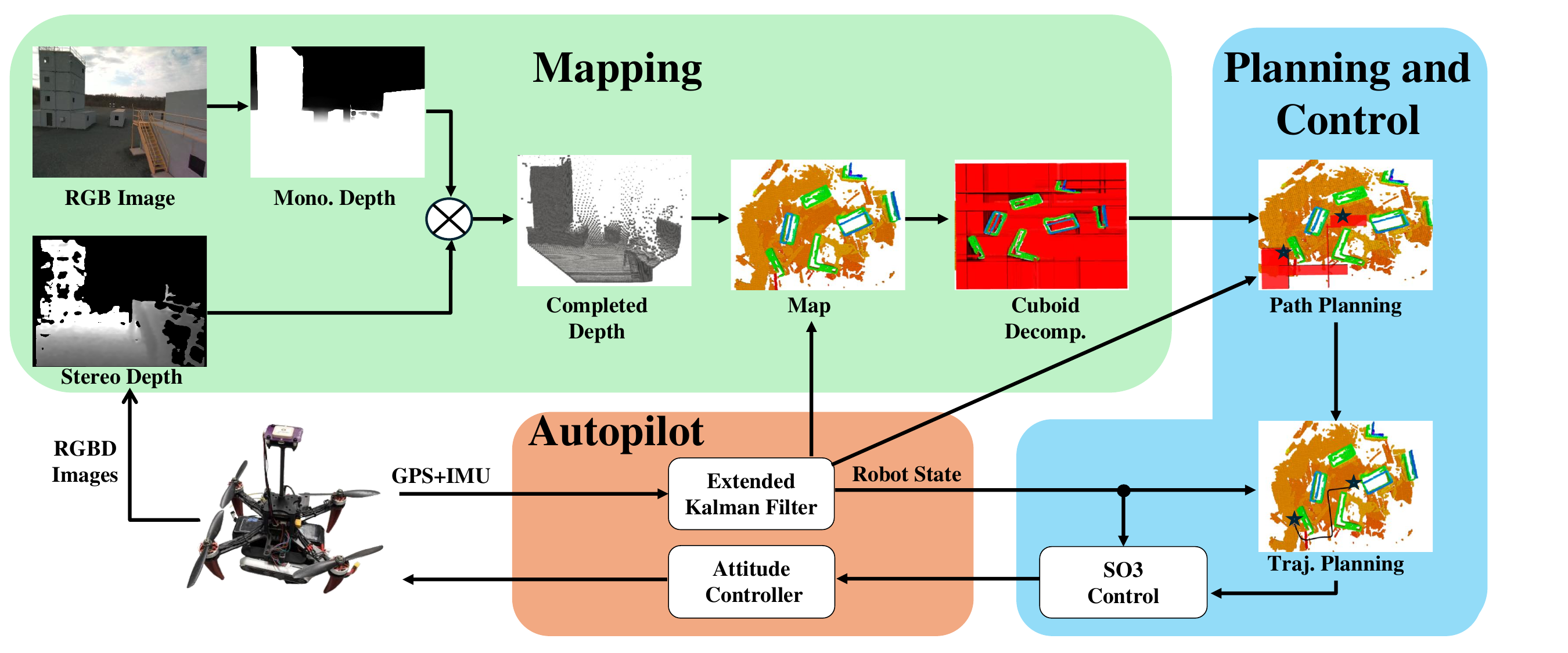}
    \vspace{-1em}
    \caption{System block diagram. The perception and planning/control pipelines are run onboard the Orin NX. Our autopilot is the Pixracer Pro running Px4 firmware. Stars represent the start and goal for the trajectory planning. The start information comes from the robot state, and the goal is user given or done autonomously with the exploration algorithm.}
    \label{fig:control_diagram}
\end{figure*}

\textbf{Planning}. There are two main planning approaches for obstacle avoidance: local planning and global planning.
Local planning \cite{saviolo2024reactivecollisionavoidancesafe,rappids, RapidQuad, PCMPC} is favored for its fast computation, making it well-suited for high-speed reactive avoidance. However, since it only considers local information, it struggles to navigate dead ends or non-convex obstacles. To handle such situations, global information is typically integrated at a higher level of abstraction through a separate global planner.
Global planning involves computing a collision-free path across the entire environment map and refining it into a feasible trajectory \cite{zju_swarm, liu2022star, liu_safe_flight_corridor, RAPTOR, Roypoly}. Global path generation methods generally fall into two categories: sampling-based and search-based approaches. 
These methods are also applicable to local planning if we restrict the map size to a local region. However, in the following, we will evaluate the performance of the following method within the context of a large-scale global map.
Sampling-based methods, such as Rapidly-exploring Random Trees (RRT) \cite{Roypoly} and the Probabilistic Roadmap Method (PRM) \cite{RAPTOR}, generate paths by randomly sampling the space and connecting traversable states to form a space filling tree. 
These methods scale well to high-dimensional spaces and avoid space discretization. However, they lack optimality guarantees and have unbounded computation times.
In contrast, search-based methods, including A* \cite{du2020multi, mohta2018fast} and jump point search \cite{liu2017searchbased}, operate on a discretized representation of the space, often uniform voxels. 
These methods provide optimal solutions, have bounded solving times, and can determine when a goal is unreachable. However, their scalability is poor in large environments as a result of the cubic growth of the search space. 
To address the poor space scalability, other works abstract the occupancy as a graph of convex sets~\cite{marcucci2024shortest,cohn2023non,marcucci2024fast}. 
The graph of convex sets representation converts the free space to a sparse set of convex objects, axis-aligned cuboids for our implementation. 
Since each element in the graph represents a full volume of uniform voxels, the number of elements in the graph is greatly decreased which simplifies the search problem.
However, previous works assume a fully known environment and use preprocessing to generate the graph. Our method is the first to apply this technique in real-time with a dynamically updating map.
Solving the path planning problem on this graph is the shortest path problem on a graph of convex sets~\cite{marcucci2024shortest,cohn2023non,marcucci2024fast}. 
Solving the path is NP-Hard however there exists a few approximation methods using either mixed integer convex programming (MICP) \cite{marcucci2024shortest} or coverts the convex sets to singular vertices \cite{marcucci2024fast} and use a shortest path on a singular graph like Djikstra, or A*. Mixed integer programming is close to optimal solution but slow to compute.  While graph based search methods are quicker to compute, but converting a set of convex sets to set of vertices is another optimization problem. 
\cite{marcucci2024fast} assumes that the map is known and solves this conversion in pre-processing.
Since our application involves real-time flight, long pre-processing steps are not feasible. To address this, our approach assumes that the all the edges of our graph of convex set are equal and constant and uses the minimum distance between each convex set and the goal as a heuristic. This design significantly improves the scalability of our path planning method in terms of computation time.

Once a set of safe flight corridors (SFC) is determined, computing a feasible trajectory becomes a well-studied problem \cite{FeiGao_opt_time, liu2022star, mohta2018fast}. One of the most popular methods is MINCO\cite{WANG2022GCOPTER, zju_swarm} which uses an unconstrained optimization with polynomial splines to plan through SFC. 
MINCO has shown better optimality in terms of time and control effort than other baselines~\cite{ Mellinger2011, gao_replan, sun_traj} in MINCO's own experimental evaluation. 
Additionally, it has show quick computation times in many of its application works~\cite{Jialin2022, ren2025safety, zju_swarm}.
We propose a constrained optimization guaranteeing our constraints are respected unlike \cite{WANG2022GCOPTER} . 
Second, we replace its polynomial spline basis with Bernstein polynomial splines \cite{bebot}. 
This change of basis  allows continuous safety enforcement without discretization~\cite{zju_swarm}, or additional post-processing for collision checks and replanning in proposed other works \cite{Roypoly, RAPTOR}. 
Additionally unlike other works using Bernstein polynomial splines \cite{bebot,bspline}, our work does not directly optimize control points or coefficients but the start and end points of each spline reducing the complexity of the optimization.
Our results show that our method takes less computation time \cite{WANG2022GCOPTER} while maintaining comparable optimality and guaranteeing safety. 

\section{Methodology} \label{sec:methodology}
In this section, we describe our navigation approach made up of two major parts as seen in Fig.~\ref{fig:control_diagram}, a mapping and planing/control pipeline. First, we require our robust mapping pipeline which generates both an occupancy map and graph of free space. 
This graph of free space is then fed into our robust planning and control algorithm design. 

\begin{table}[]
    \centering
    \begin{tabular}{c|c|c|c}
     Estimated Value & True Depth & Depth Error & Disparity Error\\
       $[1\si{m},2\si{m},5\si{m}]$  & $[1\si{m},2\si{m},10\si{m}]$ & $5\si{m}$ & $\mathbf{0.1\si{px}}$\\
       $[2\si{m},4\si{m},10\si{m}]$ & $[1\si{m},2\si{m},10\si{m}]$ & $\mathbf{2.23\si{m}}$ & $0.55\si{px}$  \\
    \end{tabular}
    \caption{Depth error versus disparity error examples. We assume $3$ objects and compare them to a ground truth depth. We see that depth error focuses on amplifies the error on the far objects. Disparity on the other hand amplifies the error on close objects. We assume a baseline of $1\si{m}$ and focal length of $1\si{m}$.}
    \label{tab:depth_example}
\end{table}
\subsection{Mapping}
In this section, we describe our depth generation method. Without accurate mapping, the following sections on planning and control are unable to be implemented in unknown environments. By solving this problem, we are able to ensure an accurate map to leverage our planning and control for safe navigation. 
We utilize an RGB-D camera that outputs temporally and spatially synchronized color and stereo disparity images. Our goal is to take the noisy disparity image created by stereo and combine it with color information to create a less noisy and structurally consistent disparity which will then be converted to depth. We prefer to do our calculations in this disparity or inverted depth scale for our application of navigation and image based sensors rather than normal depth scale. This allows us to emphasize error on the close objects where image based sensors are more accurate and is more relevant for planning. 

We will use the following example to elucidate our point. Imagine our image has $3$ main objects with a true distance of $[1\si{m},2\si{m},10\si{m}]$. 
Then let's say we have $2$ methods that give separate distance estimates. 
Method A estimates $[1\si{m},2\si{m},5\si{m}]$. Method B estimates $[2\si{m},4\si{m},10\si{m}]$. 
We see various error results in Table.~\ref{tab:depth_example}.
Depth error is computing by directly using the RMSE. Disparity error is inverting the estimated value and true depth and then calculating the RMSE.
We see that choosing disparity error amplifies the error for close range objects over the long range objects done in the depth error.
From a planning perspective, disparity error is superior because accurately representing the distance of close objects is more important than the far object. This is especially aggravated since our system is continuously replanning, and we will have less time to react to close objects as opposed to far objects when we replan. 

To perform our depth completion, we do the following steps. First, we feed our color image ($W\times H\times 3$ dimensions) is fed to a monocular depth estimation network, DepthAnythingv2 \cite{depthanything}, which generates an unscaled estimated disparity image $\mathbf{d}_m$ with units $\si{px}$. 
Monocular depth is an ill-posed problem. The structure created by the neural network is good
however the scale information is highly inaccurate. 
Conversely, stereo disparity, $\mathbf{d}_s$ in pixels or $\si{px}$, has a reliable scale but lacks structure and is filled with speckle noise and holes.
This is visible in  Fig.~\ref{fig:control_diagram} where the stereo depth and from it disparity has the several large holes in the building structures. 
Ideally, we want to fuse the structure from the monocular depth with the scale from the stereo. To achieve this objective, we rescale our monocular depth with the following optimization problem

\begin{equation}
\begin{split}
    \mathbf{d}_c[i,j] &= \alpha_2 \mathbf{d}_m[i,j]^2 + \alpha_1 \mathbf{d}_m[i,j] + \alpha_0,\\
    \min_{\alpha_{2,1,0}}  \sum_{j=0}^W\sum_{i=0}^H &  \left(\mathbf{d}_s[i,j] -  \mathbf{d}_c[i,j]\right)^2.  
\end{split}
\end{equation}
where $\mathbf{d}_c$ is the completed disparity image in $\si{px}$, $[\alpha_2,\alpha_1,\alpha_0]$ are scaling coefficients, and $i$ and $j$ are the index terms for the depth images. We see the optimization is a simple polynomial fitting using a quadratic curve. 
The scaling coefficients are solved using the least squares regression algorithm \cite{Weisberg2013-yq}. To generate the depth we use the stereo cameras baseline and focal length to convert the completed disparity image,$\mathbf{d}_c$ ,  to a completed depth image in $\si{m}$. The completed depth  and vehicle's pose is fed to a mapping algorithm, Nvblox \cite{millane2024nvblox}, to generate the occupancy map. After the occupancy is generated, we extract the free space and perform the cuboid decomposition described in the following. 

\subsection{Cuboid Decomposition and Path Selection}
To generate an obstacle-free, dynamically feasible trajectory, our approach start by generating a set of Safe Flight Corridors (SFC) to describe the map.
Most existing methods \cite{WANG2022GCOPTER, liu2022star, liu_safe_flight_corridor} first compute a obstacle free path and then create SFC around the path.
Our method instead reverses this procedure converting the occupancy map to a graph of SFC, and then finding the subset of SFC that links start and goal. 
We call our approach to turn the occupancy map into a graph of safe flight corridors cuboid decomposition.
The advantage of our approach is that the clustering can be performed at the map generation step in parallel to trajectory planning before a goal is given.
Additionally, once the clustering is complete,  solving the subset of safe flight corridors to connect the starting pose and goal is orders of magnitude faster.

\begin{algorithm}[t]
\caption{Cuboid Decomposition}
\label{alg:cuboid_decomp}
\begin{algorithmic}
\small
    \item[\textbf{Input:}] Occupancy Grid 
    \item[\textbf{Output:}] A graph of cuboids with a set of vertices $\mathbb{V}$ that represent free space sets and edges $\mathbb{E}$ that represent their connectivity
    \STATE \textbf{Initialization}: Expand the obstacles by the control error and vehicle's radius. Mark all free space as uncovered.
 \WHILE{There exists an uncovered node}
     \STATE \textbf{Node Selection}:  Select an uncovered and unoccupied node, as shown in the search step (Fig.~\ref{fig:cube_decomp}a).
     \STATE \textbf{Cover}: Expand a cuboid $v_k$ from the selected node in all directions until it touches obstacles on each side seen in Fig.~\ref{fig:cube_decomp}b.  
     \STATE \textbf{Add Vertices} $\mathbb{V} =v_k \cup \mathbb{V}$
     \STATE \textbf{Mark}: Mark all voxels within the added cuboid as covered.
 \ENDWHILE
 \STATE \textbf{Solve Edges}
\FOR{$v_i \in [\mathbb{V}]$}
\FOR{$v_j \in [\mathbb{V}]$}
\IF{$v_i \cap v_j \neq \varnothing $}
    \STATE \textbf{Make Edge} $e_k = [v_i,v_j]$\
    \STATE \textbf{Add Edge} $\mathbb{E} =e_k \cup \mathbb{E} $
\ENDIF
\ENDFOR    
\ENDFOR
\RETURN Graph of Convex Sets $\mathbb{V}$ and $\mathbb{E}$ 
\end{algorithmic}
\end{algorithm}

\begin{figure}[!t]
    \centering\includegraphics[width=\linewidth]{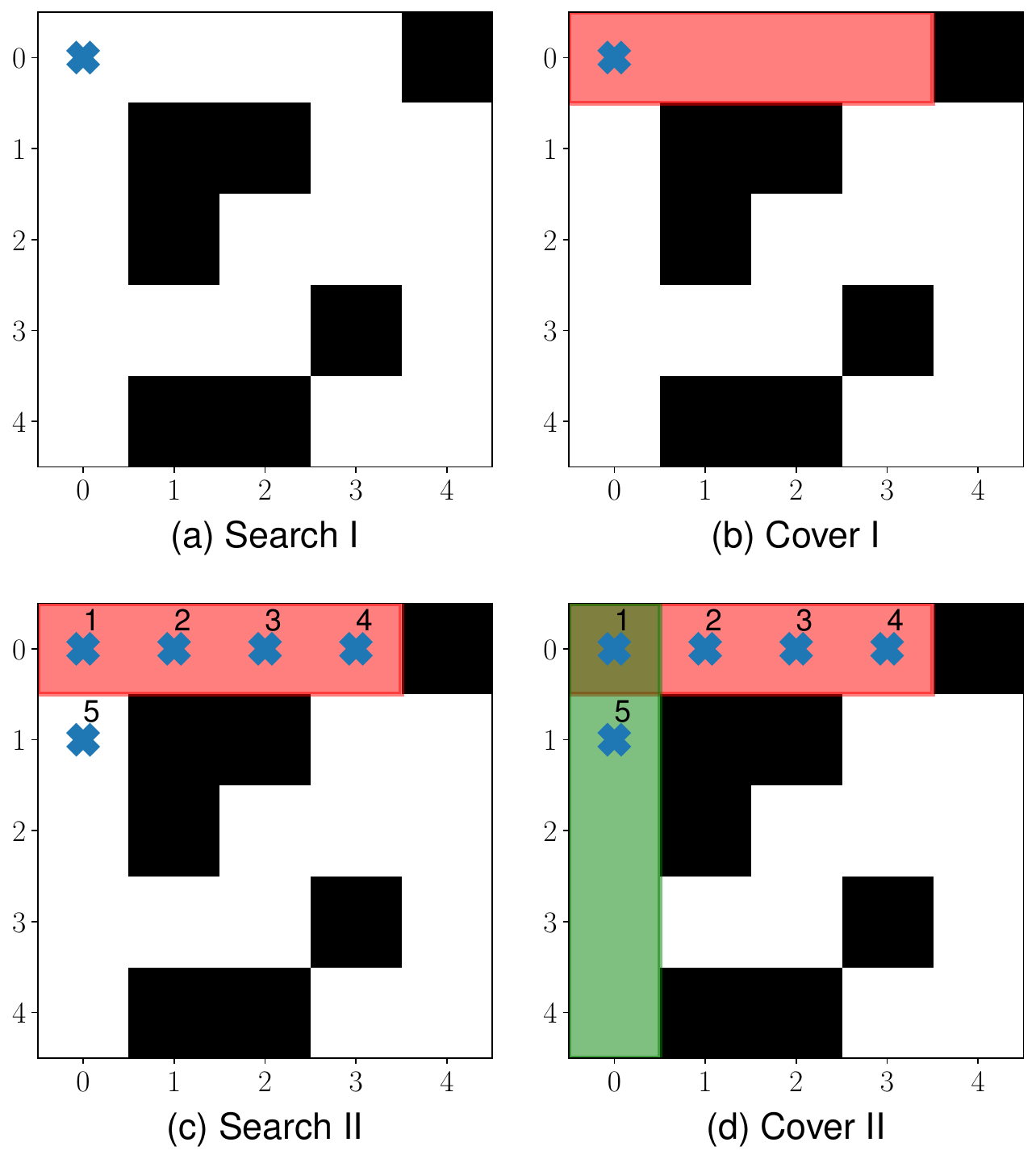}
    \vspace{-1em}
    \caption{Cuboid Decomposition:  First discover an unexplored and unoccupied node. Next create a cuboid to cover the free space expanding in all directions.}
\label{fig:cube_decomp}
\end{figure}
\begin{figure}[t]
    \centering
    \includegraphics[trim={0 1cm 0 0},clip,width=\linewidth]{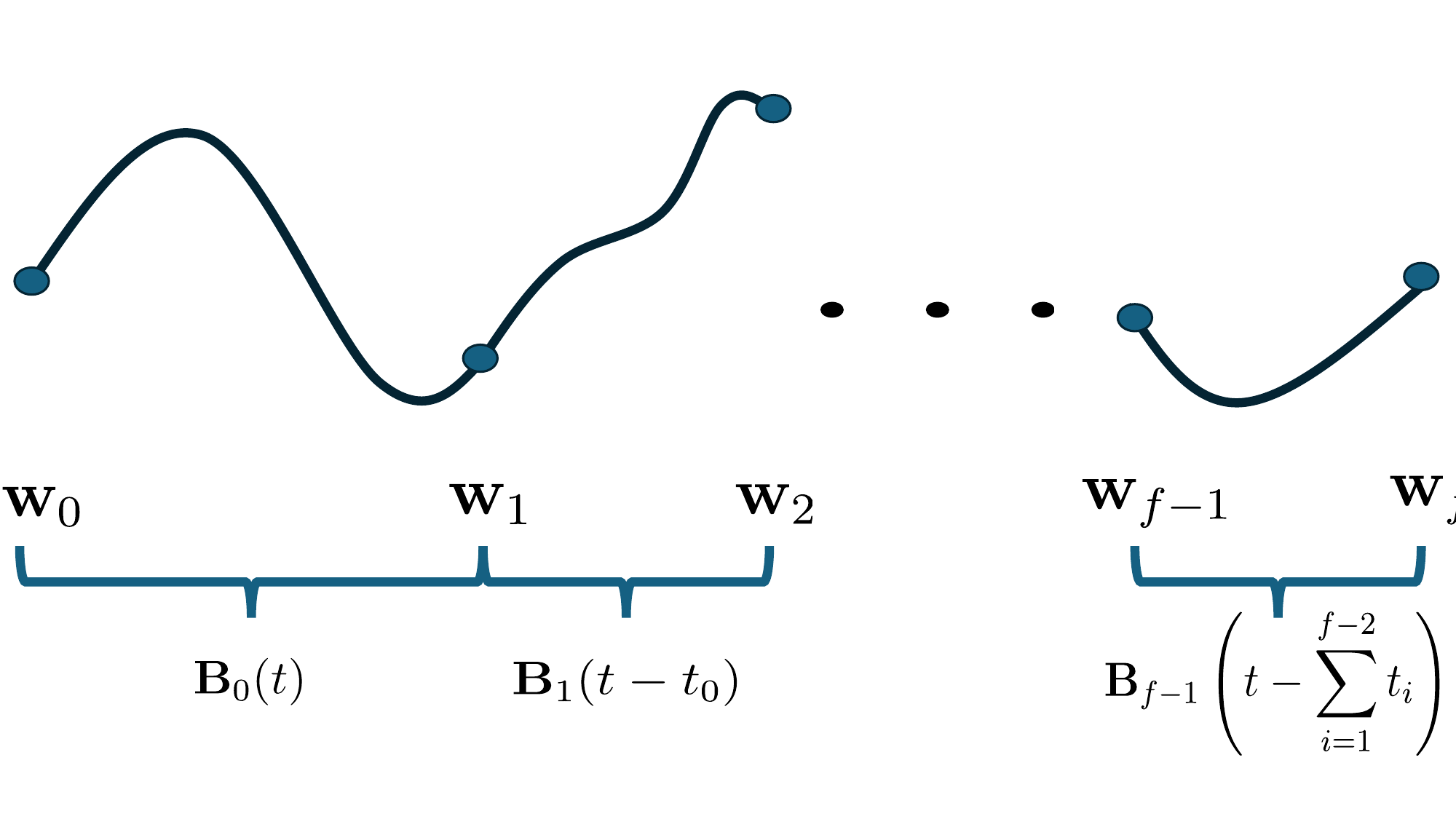}
    \vspace{-1em}
    \caption{Spline Definition: $\mathbf{w}_i$ the waypoint number. $\mathbf{B}_i$ Bernstein spline between each waypoints. }
    \label{fig:spline_defintion}
\end{figure}

In the literature, cuboid decomposition can be viewed as a covering problem, where the goal is to cover the free space with boxes, as described in \cite{FRANZBLAU1984164}. 
Our proposed cuboid decomposition algorithm is illustrated in Fig.~\ref{fig:cube_decomp}. The procedure involves the following steps described in our Algorithm$~\ref{alg:cuboid_decomp}$. We initialize the space by expanding the obstacles by the average control error, localization error, and vehicle radius.
Then we set all free space to uncovered. 
After initialization, we repeatedly create cuboids to cover the free space. Subsequently, we construct the edges of our graph by checking if each pair of cuboids share points. As a result, each vertex of this graph represents a cuboid of free space and each edge represents an adjacency. 
This graph-based representation is updated repeatedly, independently of path planning.
To extend this decomposition into $2.5$D, we divide the space into several $2$D layers and repeat the process for each layer. 
This process is done by declaring a minimum and maximum flying height for the vehicle. We then divide this minimum and maximum height equally among the $3$ layers along with some buffer height.

Once we have a graph representation of the free space, we solve for our path. Rather than an explicit path, we are selecting a set of SFC that connect our start and goal positions. We implement a modified A* algorithm to achieve our goal. 
The edge weight of each corridor is $1$. The lower bound heuristic is the minimum distance between a point in the cuboid node and the goal, assuming that no obstacles exist.  
We include in our appendix why other edge weight metrics such as centers of the convex sets or minimum distance between sets is less optimal than a constant edge weight. 
As the number of cuboids is significantly lower than the nodes in an occupancy grid, A* can be solved in a fraction of time while providing a near-optimal set of safe flight corridors as seen in Fig.~\ref{fig:traj_planning}c. Solving for a true optimal set of safe flight corridors is a NP-Hard problem \cite{marcucci2024fast}. Conversely, this algorithm's objective is to minimize the number of SFC which in turn corresponds to minimizing the number of optimization variables for our trajectory generation.

\subsection{Bernstein Time-Optimized Trajectory Planning}
\begin{figure*}[t]
        \centering\includegraphics[width=\linewidth]{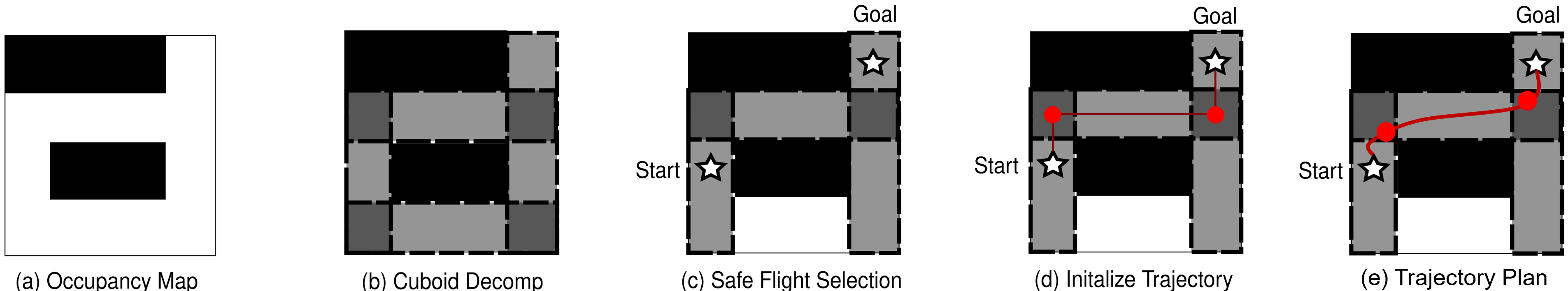}
        \vspace{-1em}
    \caption{Trajectory Planning Pipeline. Cuboid Decomposition refers to the procedure detailed in Fig.~\ref{fig:cube_decomp} and Algorithm~\ref{alg:cuboid_decomp}. Safe Flight Selection refers to finding the minimum number of safe flight corridors that connect a start and a goal. The light gray area is where the cuboids are covering the free space. Darker gray areas are just where the safe flight corridors intersect. Red dots are the optimization variable which are initialized in step (d) and finally optimized in step (e).}
    \label{fig:traj_planning}
\end{figure*}
We solve for a trajectory that connects a start condition of position, velocity, and acceleration $[\mathbf{x}_0^\top, \mathbf{v}_0^\top, \mathbf{a}_0^\top]^\top$ with a goal position $\mathbf{x}_f$ given a set of safe flight corridors.
We introduce the formalism for a single-dimensional case. To extend this to multiple dimensions, it is sufficient to repeat the constraints for each dimension. We formulate our trajectory using a fifth order Bernstein polynomial splines. Each spline is assigned to a single safe flight corridor.
First, we define each spline as follows
\begin{equation} \label{eqn:poly_spline}
\begin{split}
B(t) &= 
     \begin{cases}
       \text{$B_{0}\left(t\right)$} &\quad\text{if $t 	\in \left[0,t_0\right]$}\\
        \text{$B_{1}\left(t-t_{0}\right)$}&\quad\text{if $t 	\in \left[t_{0},t_0+t_{1}\right]$}\\
        \hspace{10pt}\vdots\\
        \text{$B_{f-1}\left(t-\sum\limits_{i=1}^{f-2}t_i \right)$} &\quad\text{if $t 	\in \left[\sum\limits_{i=1}^{f-2}t_i ,\sum\limits_{i=1}^{f-1}t_i \right]$}\\
     \end{cases},\\
\end{split}
\end{equation}
where $t_i$ represents the time spent in the safe flight corridor $i$. $f$ is the number of safe flight corridors.  
Each spline $B_i(t)$ is defined as the following function
\begin{equation} \label{eqn:basis_def}
\begin{split}
B_i(t) &= \sum_{k=0}^5 c_{k,i}  b_{k,5}\left(\frac{t}{t_i}\right),\\
\end{split}
\end{equation}
where $c_{k,i}$  is the coefficient of the polynomial, $k$ represents the coefficient number and $i$ is the spline number. In literature, an alternative name for the coefficient is control points when describing Bernstein polynomials. $b_{k,5}(t)$ represents the basis function defined in \cite{bebot} for order $5$  and 1-D polynomial. We can see in fig.~\ref{fig:spline_defintion} a visualization of our spline. 

However instead of the optimization problem to solve for coefficients, we structure our optimization as solving for a set of waypoints $\mathbf{w} = [\mathbf{w}_0, \mathbf{w}_1,..., \mathbf{w}_{f}]^\top$ and traversal times $\mathbf{t} = [t_0, t_1,...t_{f-1}]^\top$ . 
We define a waypoints as a $\mathbf{w}_i \in \mathbb{R}^{3\times 1}$ with elements $\mathbf{w}_i = \begin{bmatrix}
    B_i(0) & B_i'(0) & B_i''(0) 
\end{bmatrix}^\top$ where $B' = \frac{dB(t)}{dt}$ is the velocity, and $B''= \frac{d^2B(t)}{dt^2}$ is acceleration. These waypoints are visualized in fig.~\ref{fig:spline_defintion}. In our formulation, we have $f$ splines and $f+1$ waypoints describing these splines.
Each spline is completely defined by the start and end waypoint on the curve. This relationship between coefficients and waypoints is defined as 
\begin{equation}\label{eqn:way_coeff_convert}
\begin{bmatrix}
c_{0,i}\\
c_{1,i} \\
\vdots \\
c_{4,i}\\
c_{5,i}\\
\end{bmatrix}
 = \begin{bmatrix}
1 & 0 & 0 & 0 &0 & 0\\
1 & \frac{1}{5}t_i & 0 &  0 & 0  & 0\\
1 & \frac{2}{5}t_i & \frac{1}{20}(t_i)^2  & 0 &  0 & 0 \\
0 & 0 & 0 & 1& -\frac{2}{5}t_i & \frac{1}{20}(t_i)^2 \\
0 & 0 & 0 & 1 & -\frac{1}{5}t_i & 0  \\
0  & 0 &  0 & 1 & 0 & 0\\
\end{bmatrix}
\begin{bmatrix}
\mathbf{w}_{i} \\
\mathbf{w}_{i+1} \\
\end{bmatrix}
,
\end{equation}

The derivation of eq.~\ref{eqn:way_coeff_convert} is included in our appendix. Building on this structure, we formulate the trajectory generation problem with the following optimization 
\begin{equation}\label{eqn:traj_op_problem}
\begin{aligned}
\min_{\mathbf{w}_{[0,...,f]}, \mathbf{t}_{[0,...,f-1]}} \quad &  ~\sum^{f-1}_{i=0}t_i,\\
\textrm{s.t.} \quad \mathbf{w}_0&=\begin{bmatrix}
    \mathbf{x}_0^\top & \mathbf{v}_0^\top & \mathbf{a}_0^\top
\end{bmatrix}^\top,\\
\mathbf{w}_{f}&=\begin{bmatrix}
    \mathbf{x}_f^\top & 0 & 0
\end{bmatrix}^\top,\\
 \mathbf{y} \leq &\mathbf{G}(\mathbf{t}) \mathbf{w} \leq \mathbf{z},~ 0 < \mathbf{t}    \\
\end{aligned}
\end{equation}

 The matrices  $ \mathbf{y},\mathbf{G}, \mathbf{z}$ are safety constraints based on eq.~(\ref{eqn:ineq_const}). $ \begin{bmatrix}
    \mathbf{x}_0^\top & \mathbf{v}_0^\top & \mathbf{a}_0^\top
\end{bmatrix}^\top$ is the quadrotor's current position, velocity, and acceleration respectively, $ \mathbf{x}_f$ is the goal position.  This results in a time-optimized trajectory\ in Fig.~\ref{fig:traj_planning}e.

Structuring the problem using waypoints instead of coefficients improves our optimization in three ways. First, optimizing on waypoints reduces the number of optimization variables compared to coefficients when increasing the number of splines. For example, with $f$ number of splines with $6$ coefficients and $1$ time variable per spline, we would have $7f$ optimization variables. Using our method, the number of optimization variable is $3(f+1)+f-6$ or  $4f-3$ optimization variables. We have $f+1$ waypoints, $f$ time variables, and since the start and end waypoints are set to values we can replace them with constants resulting in $-6$ optimization variables. Next, the gradients are much better conditioned on functions like $t_i^2$ and $t_i$ as opposed to optimizing over coefficients which involve the inverse, $\frac{1}{t_i}$ or $\frac{1}{{t_i}^2}$ like in eq.~\ref{eqn:coeff_spline}. Finally, it is easier to initialize our optimizer because waypoints have a better interpetability than coefficients. 

Subsequently, we define our safety constraints. The safe flight corridors are defined by their position constraints lower $\mathbf{l} = [l_0,l_1,...l_{f-1}]^\top$ and upper $\mathbf{u} = [u_0,u_1,...u_{f-1}]^\top$ for each corridor.  With these bounds, we formulate our safety constraints from eq.~(\ref{eqn:traj_op_problem}), $\mathbf{y}$, $\mathbf{G}, \mathbf{z}$, based on the convex hull property of Bernstein polynomials \cite{bebot}. We prove this property in the appendix of this work as well. The convex hull property states bounds on the coefficients are equivalent to bounds on the full trajectory defined formally as 

\begin{equation} \label{eqn:ineq_const}
\begin{split}
  l_i \leq  \mathbf{A}_i\begin{bmatrix}
\mathbf{w}_{i} \\
\mathbf{w}_{i+1} \\
\end{bmatrix} \leq u_i &\implies 
 l_i \leq B_i(t)  \leq u_i \hspace{5pt},\\
  -\mathbf{v}_m \leq  \mathbf{D}^{(1)}_i\mathbf{A}_i\begin{bmatrix}
\mathbf{w}_{i} \\
\mathbf{w}_{i+1} \\
\end{bmatrix} \leq  \mathbf{v}_m &\implies 
 -\mathbf{v}_m \leq B'_i(t)  \leq \mathbf{v}_m ,\\
  \underbrace{-\mathbf{a}_m \leq  \mathbf{D}_i^{(2!)}\mathbf{A}_i\begin{bmatrix}
\mathbf{w}_{i} \\
\mathbf{w}_{i+1} \\
\end{bmatrix}\leq \mathbf{a}_m}_{\text{Optimization Constraint}}&\implies 
 \underbrace{-\mathbf{a}_m \leq B''_i(t) \leq \mathbf{a}_m}_{\text{Physical Result}}   ,\\
\end{split}
\end{equation}
where $\mathbf{v}_m$, $\mathbf{a}_m$ is the maximum velocity and acceleration respectively.  $\mathbf{A}_i$ is the matrix to map waypoints to coefficient defined in eq.~(\ref{eqn:way_coeff_convert}). $\mathbf{D}_i^{(m!)}\in \mathbb{R}^{ (6-m) \times 6}$ is a differentiation matrix that converts the coefficient to the $m$-th time derivative's coefficient. The matrix  $\mathbf{D}_i^{(m!)}$ is composed from $\mathbf{D}_i^{(m)} \in \mathbb{R}^{ (6-m) \times (7-m)}$ as
\begin{equation}\label{eqn:bern_deriv}  \begin{split}
    \mathbf{D}^{(m)}_i &= \frac{6-m}{t_{i}}
\begin{bmatrix}
 -1 & 1 & 0 &  0 & \hdots  \\
 0 & -1 & 1 & 0 &  \hdots \\
  &  &   \ddots & \ddots  & \\
\end{bmatrix},  \\
\mathbf{D}^{(m!)}_i &= \mathbf{D}^{(1)}_{i} \mathbf{D}^{(2)}_{i} \hdots  \displaystyle 
\mathbf{D}^{(m)}_{i}.
\end{split}
\end{equation}
This matrix derivation in eq.~\ref{eqn:bern_deriv} is also included in our appendix. We then generate the initial trajectory $\mathbf{w}_{\text{init}}$. This trajectory is a straight line trajectory between the start point and the center of the intersections of each safe flight corridors visualized in Fig.~\ref{fig:traj_planning}d or mathematically defined as

 \begin{equation}\label{eqn:init_conditions}
 \begin{split}
        \mathbf{x}_i &= \frac{1}{2}(\max     [l_i,l_{i+1}]  +\min [u_i,u_{i+1}]), \\
        \mathbf{w}_i &=\begin{bmatrix}
         \mathbf{x}_i^\top & 0 & 0  \\
     \end{bmatrix}^\top,  \hspace{10pt} 1 \leq i \leq f-1, \\ 
         \mathbf{w}_{\text{init}} &= \begin{bmatrix}
         \mathbf{w}_0^\top & \mathbf{w}_1^\top
        &\hdots &\mathbf{w}_i^\top &  \hdots & \mathbf{w}_{f}^\top\\
     \end{bmatrix}^\top, \\
         \mathbf{t}_{\text{init}} &= \begin{bmatrix}
            \frac{||\mathbf{x}_1- \mathbf{x_0}||}{||\mathbf{v}_m||}  &\frac{||\mathbf{x}_2- \mathbf{x_1}||}{||\mathbf{v}_m||} & \hdots & 
            \end{bmatrix}^\top.
 \end{split}
 \end{equation}
After we acquire our initialization, we complete the optimization described in eq.~(\ref{eqn:traj_op_problem}) to solve for the the time-optimized trajectory. Since our camera has a limited FoV and is forward facing, we need to constantly align our camera with the axis of motion.  We implement a simple planner for determining the $\psi_d$ or desired yaw of the vehicle. 
\begin{equation}
\psi_d(t) = 
     \begin{cases}
       \psi_c &\quad\text{if $ ||v_y||, ||v_x|| < 1e-1 $}\\
        \psi_c-\psi_m&\quad\text{if $\arctan(v_y,v_x) - \psi_c  < -\psi_m$}\\
        \psi_c+\psi_m&\quad\text{if $\arctan(v_y,v_x) - \psi_c  > \psi_m$}\\
        \arctan(v_y,v_x)&\quad\text{else}
     \end{cases}
\end{equation}
where $\psi_c$ is the current yaw angle. $\psi_m$ is a limit on the maximum turning speed. $v_y$ and $v_x$ are the quadrotors current velocity projected to the world's $x$ and $y$ axes.

To handle newly observed obstacles, we implement a repeated replanning procedure. First, we estimate if our currently planned trajectory intersects with the updated map's obstacles within the next $10~\si{s}$. If no collision is detected, then we continue with this flight plan. Otherwise, we plan a new trajectory with the updated map to our goal. In the case where no new trajectory can be solved, we stop the quadrotor (for example the goal is in a wall). This check of deciding whether to continue, stop, or replan is repeated at a rate of $20~\si{Hz}$. 

\begin{figure*}[t!] 
\centering\includegraphics[width=\linewidth]{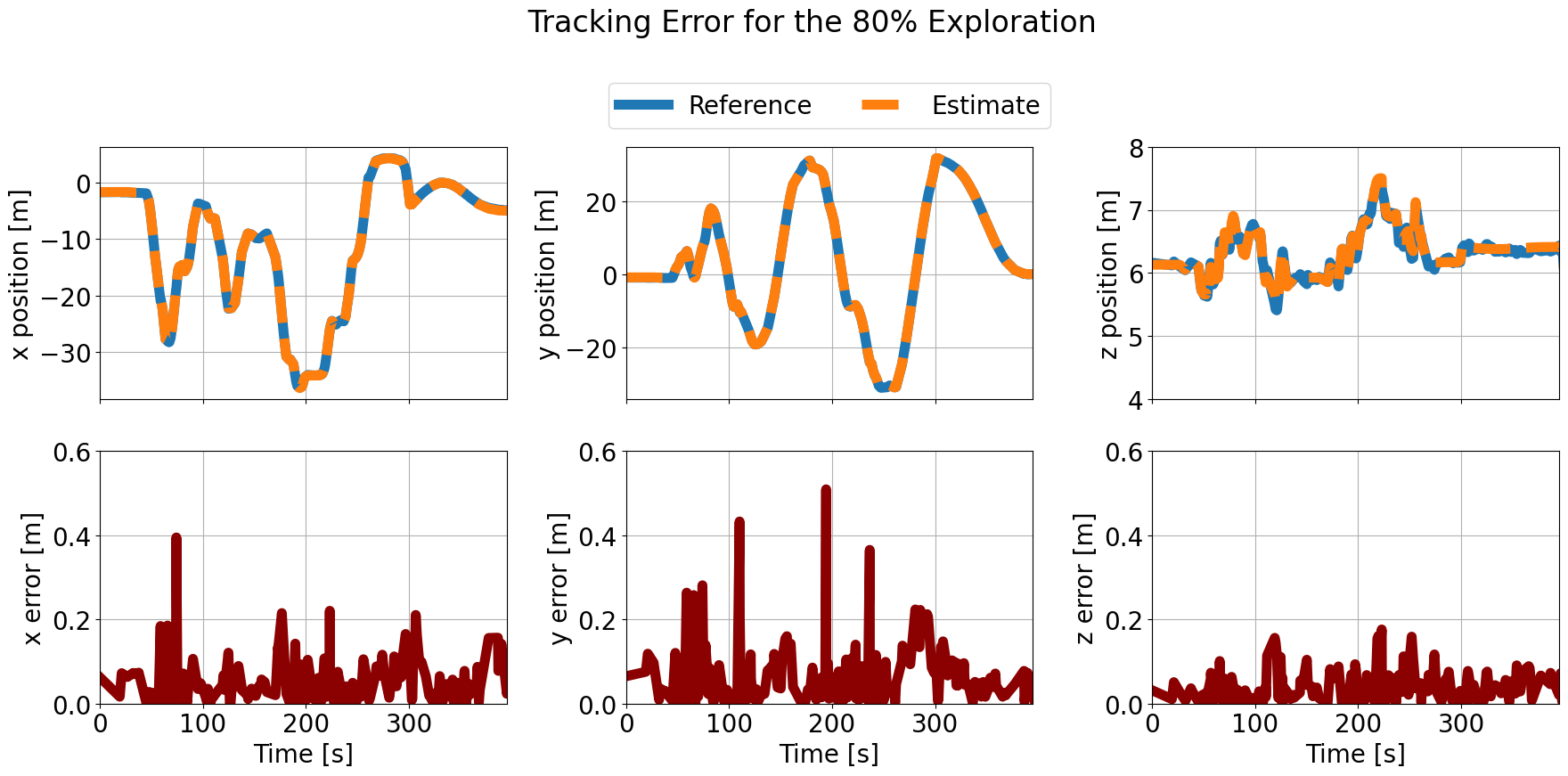}
    \vspace{-1em}
    \caption{The reference and estimate path of the $80\%$ exploration run including the tracking error.}
    \label{fig:tracking_error}    
\end{figure*}   

\begin{figure}[t!]    \centering\includegraphics[width=\linewidth]{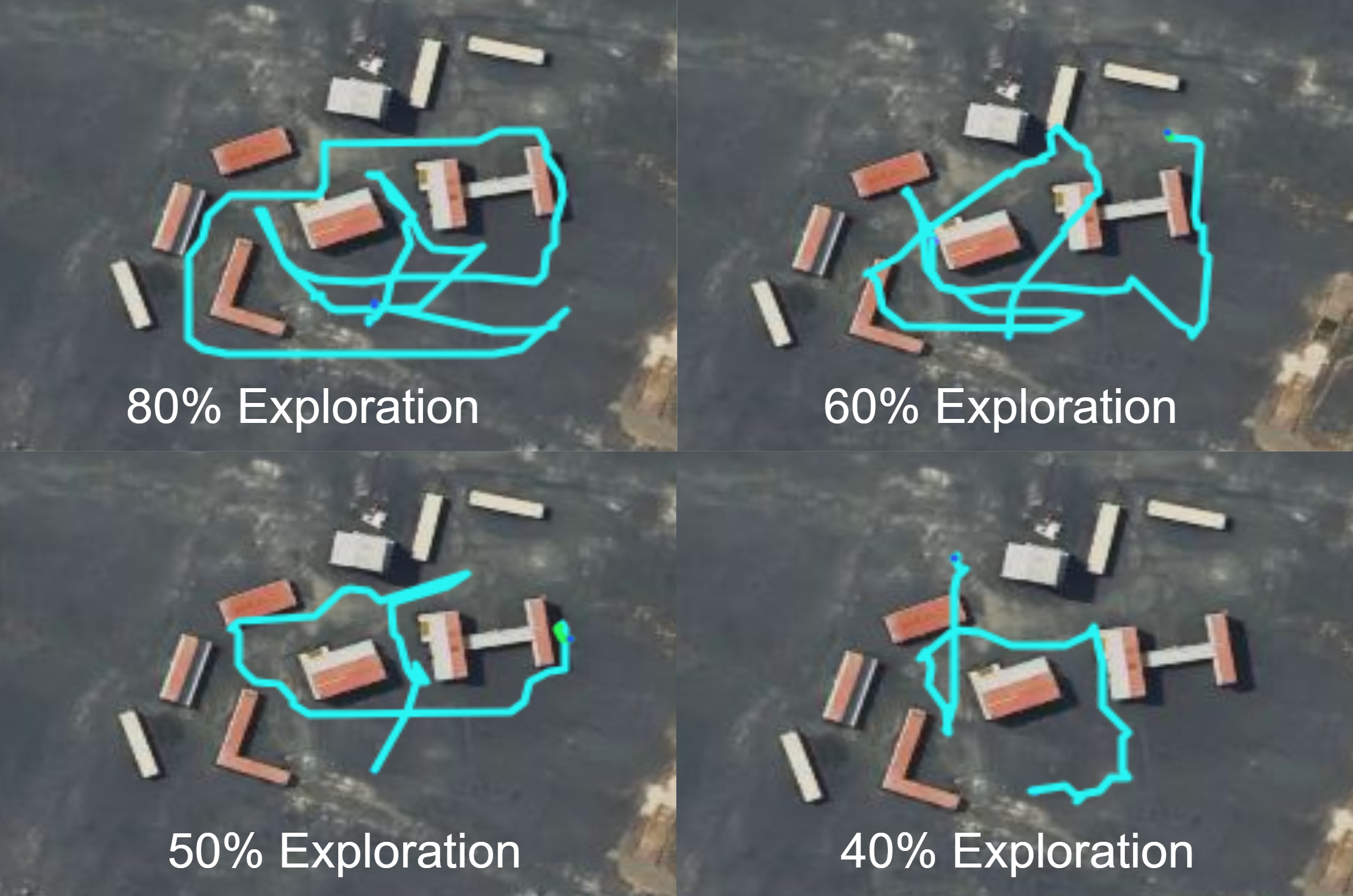}
    \vspace{-1em}
    \caption{Outdoor Exploration: We show our exploration paths given different completion goals in the environment.}
    \label{fig:explr_path}
\end{figure}

\begin{figure}[t!]    \centering\includegraphics[width=\linewidth]{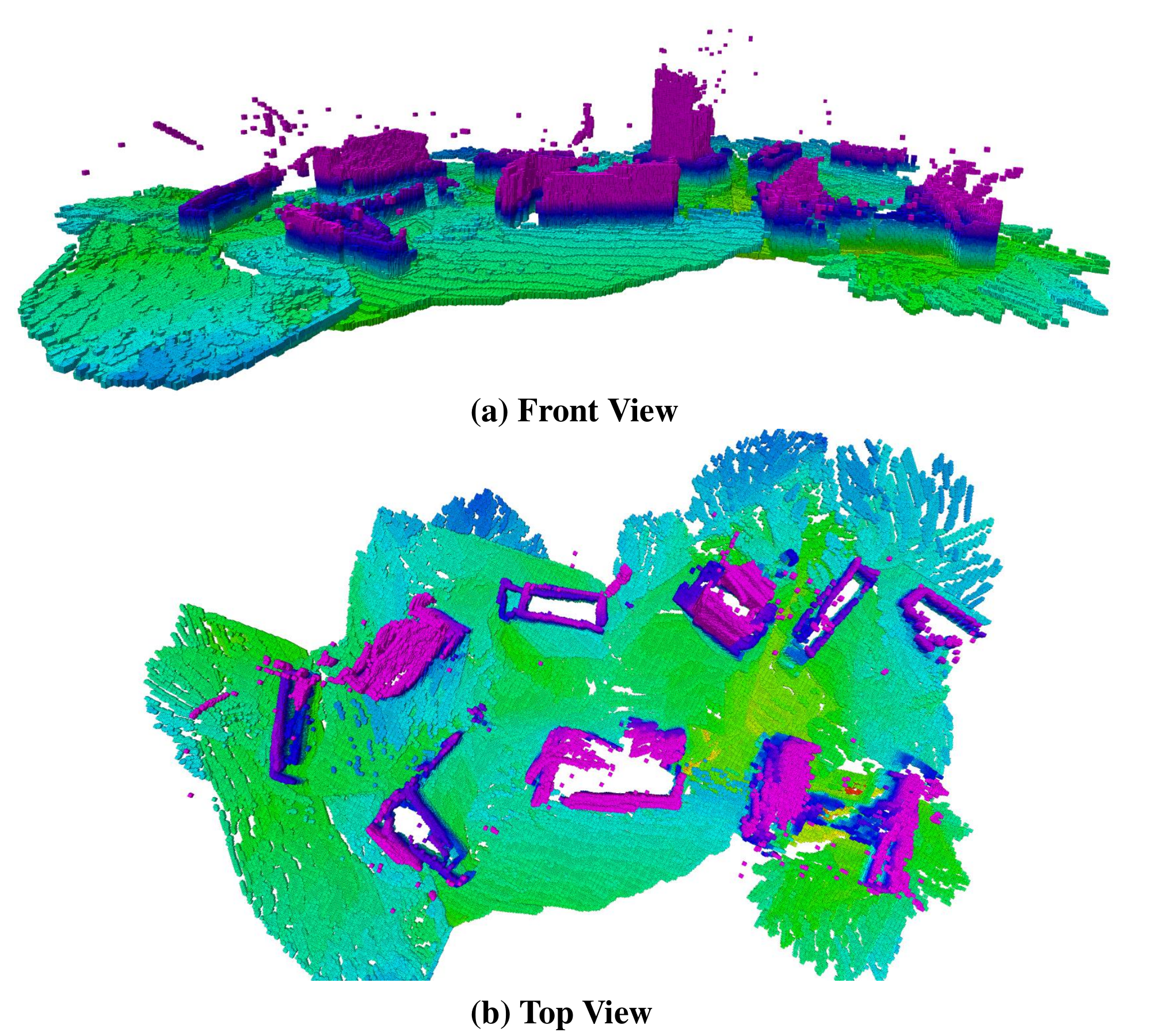}
    \vspace{-1em}
    \caption{The generated map by the quadrotor after an exploration run.}
    \label{fig:explr_map}
\end{figure}

\begin{figure*}[t]    
    \centering
    \hspace*{-1.25cm}
    \includegraphics[width=\linewidth]{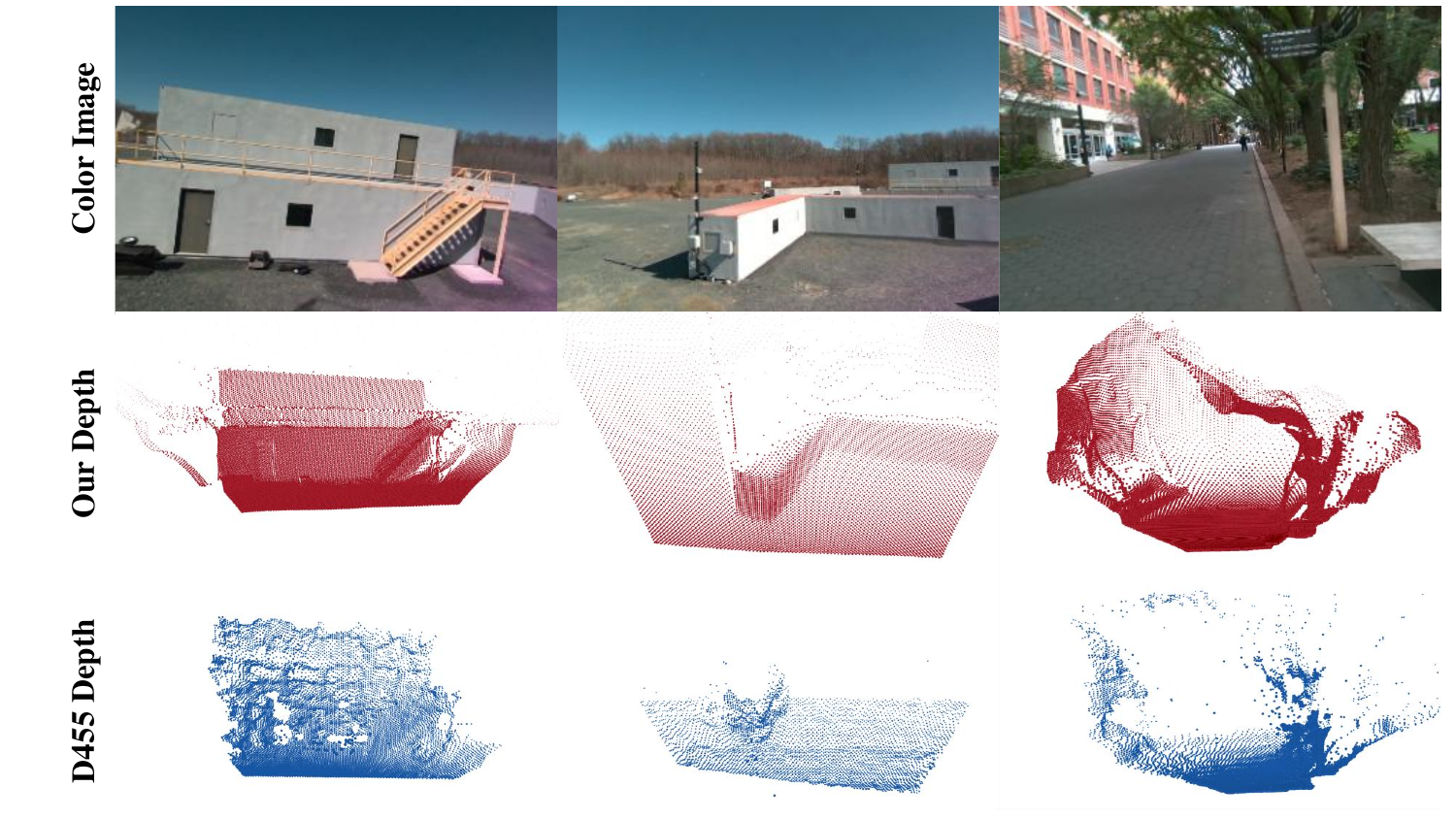}
    \caption{Depth Comparison. D455 Depth in blue is the stereo depth generated directly from the Intel Realsense D455 Depth Camera. This is a classical computer vision algorithm similar to block matching. Our Depth is the depth completion method described in our methodology. We see performing our method already gives superior results than the consumer grade depth camera. }
    \label{fig:depth_comparison}
\end{figure*}

\subsection{Exploration}
With our navigation pipeline, exploration is framed as goal selection. As exploration is not our primary contribution and is instead an application for navigation, we employ the Next Best View (NBV) algorithm \cite{NextBestViews} for its simplicity.
First, the user defines a space to explore the environment. 
Within this space, we maintain a discretized voxel map representing whether this position has been viewed. Initially, the entire map is marked as unexplored. 
As the camera observes the environment, visible voxels are updated to be explored, gradually constructing an understanding of the area.
To start the exploration, the robot performs a $360^\circ$ scan at the beginning of each mission.
This maneuver provides initial information for our exploration and obstacle avoidance. 
Goal selection is handled using a sampling-based approach called NBV.
We randomly sample $300$ candidate viewpoints within the voxel grid and evaluate how many unexplored voxels each viewpoint would newly reveal. 
We then select the viewpoint that maximizes this expected information gain. Navigation proceeds toward the selected goal until the goal is reached or found unreachable. 
Afterward, the quadrotor stops and selects another goal to continue the exploration. 
This exploration loop continues until an user discovery threshold is reached.

\section{Experimental Results} \label{sec:Results}
For our experiments, we want to show our navigation framework is effective in both indoor and outdoor scenarios, and that the individual components compare well to standard baselines. First, we stress our system stack by continuously exploring an outdoor arena to different completion levels. Next, we compare the trajectory generation to MINCO\cite{WANG2022GCOPTER}, a popular baseline. Our method achieves greater computational efficiency while producing trajectories that are comparable in quality to those generated by this approach.
Third, we show our depth generation method is superior to the depth output from the stereo baseline. These improvements include less noise, better structure, and a deeper range. 
Finally, our approach is capable of generalizes to indoor GPS-denied environments by trying more scenarios inside and challenging navigation tasks where the quadrotor has to escape a dead-end

\subsection{System Setup} \label{sec:exp_setup}
For our experiments, we consider a $1.3~\si{kg}$ drone with a thrust to weight ratio of $4-1$. The onboard computer is an NVIDIA Orin NX communicating with a Pixracer Pro flight controller running PX4 \cite{Px4}. All computation is performed onboard the system, and the station only handles visualization and simple initialization commands.  This software stack is integrated with ROS2. Our sensing module is an Intel D455 RGB-D Camera. 
For localization we run either visual-inertial odometry algorithm OpenVINS \cite{openvins} fused with IMU through an Unscented Kalman Filter (UKF) for indoor flight or GPS module for outdoor flight. 
Our monocular depth estimation network is DepthAnythingv2\cite{depthanything}. 
This model is optimized using TensorRT to ensure high speed inference at $80~\si{Hz}$ inference time onboard.
To solve our trajectory optimization in eq.~(\ref{eqn:traj_op_problem}), we use Interior Point Optimizer (IPOPT) \cite{IPOPT} with MA57 \cite{MA57} as the linear solver.  
We set a maximum velocity of $2~\si{m/s}$ per axis and similarly a max acceleration of $4~\si{m/s^2}$ outdoors. 
This framework uses a SO(3) controller \cite{loianno_est} to covert the optimized trajectory to rotation attitude commands for our quadrotor to fly through. During our exploration experiments, we use the Next Best View\cite{NextBestViews} algorithm to stress our navigation pipeline. Our Next Best View algorithm selects the goals with sample points sampled evenly across the entire map known and unknown regions. We conduct our tests in two areas an indoor flying space with a space of $8~\si{m} \times 10~\si{m}$, or a larger outdoor flying space $60~\si{m} \times 40~\si{m}$.

\subsection{Outdoor Exploration}

\begin{figure*}[t!]
  \centering
  \includegraphics[width = \linewidth]{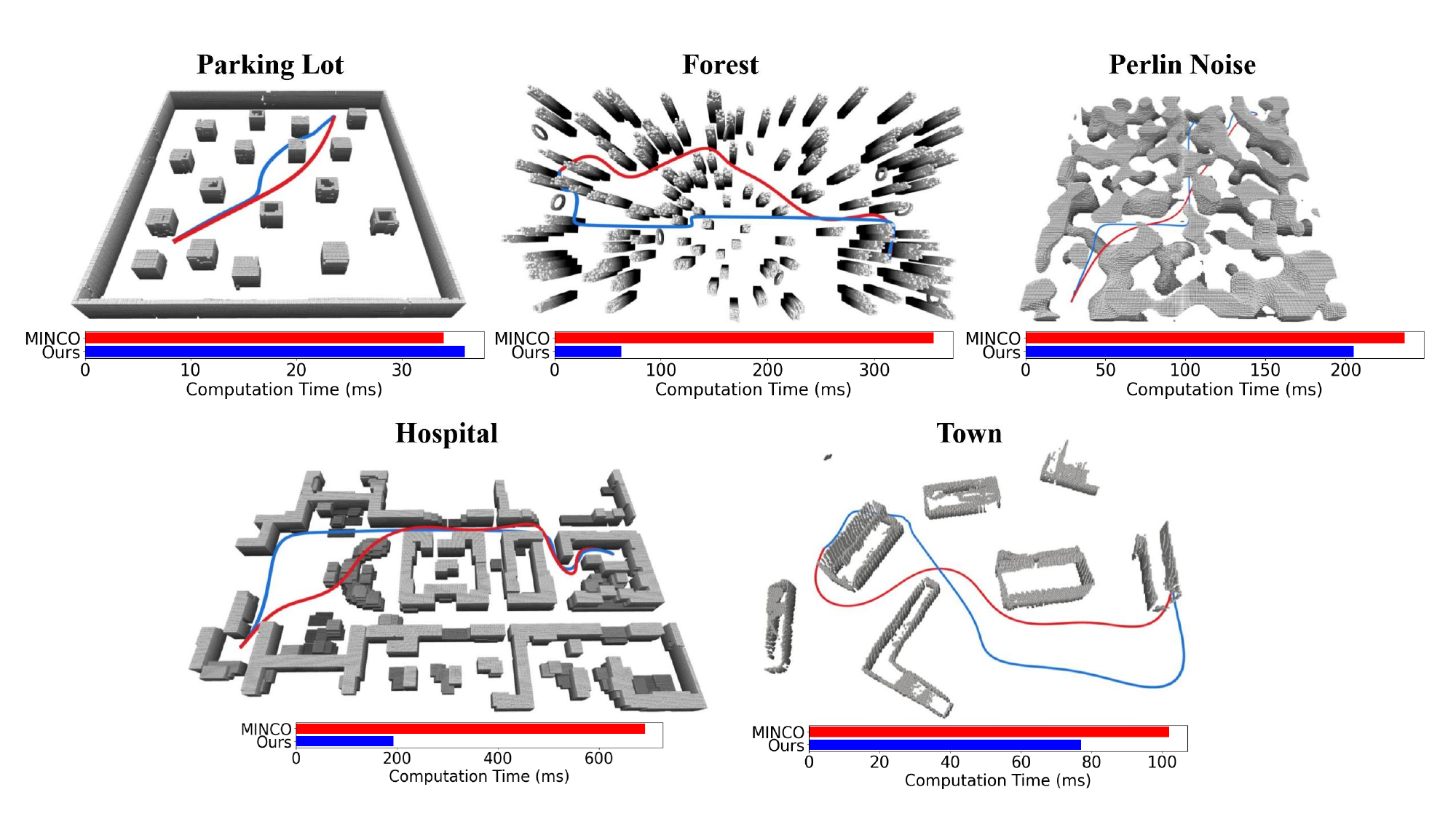}  
  \vspace{-0.5cm}
  \caption{Trajectory Path Comparison. Perlin Noise  \cite{Perlin_noise} is a gradient noise popular for procedural generation. We took the Perlin Noise Map generator from \cite{WANG2022GCOPTER}. Forest Map was borrowed from \cite{RAPTOR}. The Parking Lot map came from \cite{xu2023vision}. Hospital was taken from the Amazon Web Services Robomaker.}
  \label{fig:trajectory_path_comparison}
\end{figure*}


\begin{figure*}[t!]
  \centering
  \includegraphics[width = \linewidth]{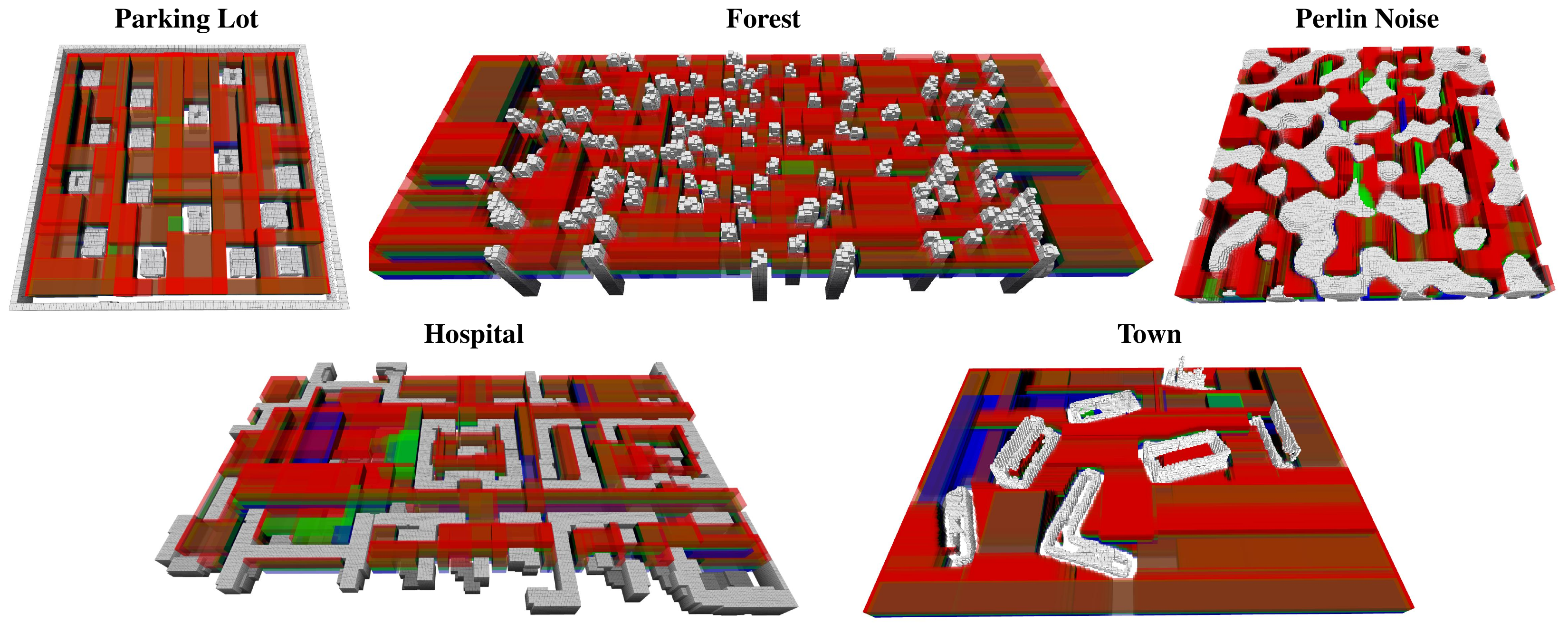}  
  \vspace{-0.5cm}
  \caption{Safe flight corridor generation. The height of these layers were manually set to prevent flying over obstacles. The colors represent the various layers of the cuboid decomposition. Red color is the top layer. Green is the middle layer. Blue is the bottom layer}
  \label{fig:cuboid_decomp_result}
\end{figure*}

In this set of experiments, we give the quadrotor exploration goal limits to reach $40\%-80\%$ of the declared volume seen in the flight path in Fig.~\ref{fig:explr_path}. From no map, the quadrotor autonomously explores a region of $60~\si{m}\times40~\si{m}$. 
Overall, we observe that as the planner is tasked with exploring a greater portion of the map, it naturally expands its search path, resulting in a more comprehensive coverage of the environment.
Additionally, the vehicle is able to safely navigate in a collision-free manner through the environment in multiple trials showing the repeatability of of the proposed solution. 
When we combine all the trial flight times, they have a cumulative $20~\si{min}$ of safe collision-free navigation in this environment.
It took around $400~\si{s}$ for the quadrotor to explore $80\%$ of the environment and about $300~\si{s}$ to explore $40\%$.
We can analyze the map generated when achieving $80\%$ exploration of the overall space based on its quality in Fig.~\ref{fig:explr_map}. The map contains all buildings and maintains a similar geometric shape. Additionally, while we observe the presence of some speckle noise in the air, this amount remains minimal.  Our tracking error and position following are close to the quadrotor desired value maintaining a tracking error of around $0.2~\si{m}$. Although some positional error spikes are observed, they are primarily attributed to environmental factors, as the weather on the testing day included an average wind speed of $15~\si{mph}$ with strong $50~\si{mph}$ gusts.

\subsection{Depth Generation Comparison}

\begin{table*}[t]
    \centering\resizebox{2\columnwidth}{!}{%
    \begin{tabular}{c|c|c|c|c}
        \hline
       Method Name  &  Method Type & Residential KITTI (RMSE)& Town KITTI (RMSE) & Scene Flow Kitti (RMSE) \\
       \hline
       Stereo Block Matching & Stereo Classical Vision &  $0.062$ & $0.034$ &  $0.050$  \\
       Metric DepthAnything & Monocular NN &  $0.57$ & $0.21$ &  $0.12$  \\
       Our Method Linear  & Hybrid & $0.056$ & $0.030$ & $\mathbf{0.012}$ \\
       Our Method Quadratic & Hybrid & $\mathbf{0.053}$ & $\mathbf{0.029}$ & $0.013$\\
       Our Method Cubic &  Hybrid & $0.054$ & $0.030$ & $0.013$ \\
       \hline
    \end{tabular}}
    \caption{Depth Completion Comparison on Kitti Dataset\cite{kitti_dataset}. We use the inverted depth RMSE for our error metric. All error values are in  $\si{px}$ using the disparity RMSE with normalized focal length and baseline. 
    Stereo Block Matching is a standard disparity method used in many off the shelf stereo cameras. Depth Anything Metric is the pure learning monocular depth method using image transformer which adds an additional linear head to rescale the image to a depth image. Hybrid refers to our combination of image transformer scaled with classical computer vision. Our Method linear, square and cube represents polynomial fitting for order $1$, $2$, and $3$ polynomials respectively. For our images, we use Our Method *. The bold number is the minimum RMSE in the column.}
    \label{tab:depth_comp_rmse}
\end{table*}

\begin{figure}
    \centering
    \includegraphics[width=\linewidth]{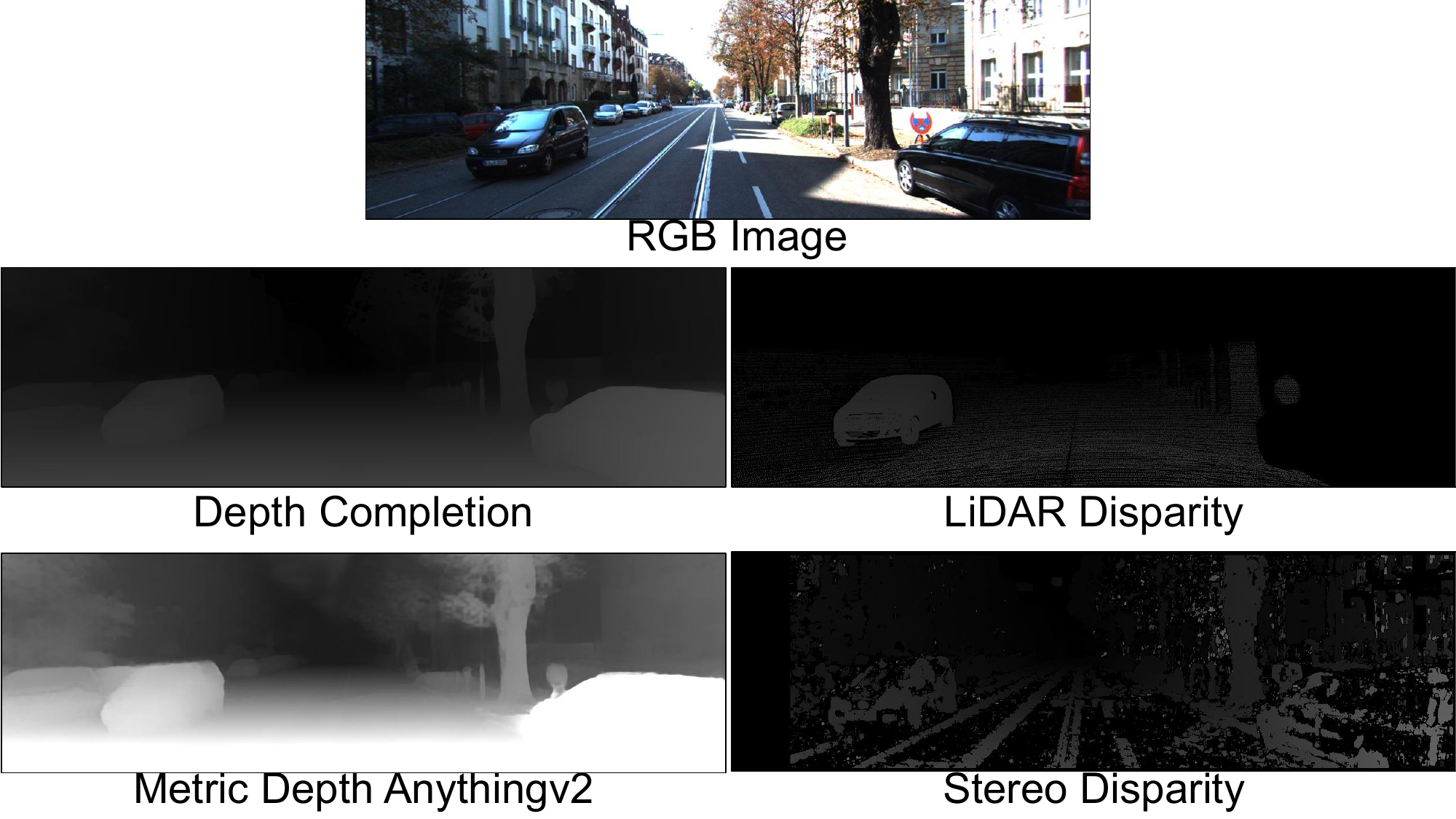}
    \caption{The top image is the color input from the left camera of the stereo pair. The bottom four images show disparity maps produced by different methods. LiDAR provides the ground truth, obtained by projecting LiDAR depth measurements into the left camera frame. Depth Completion is our hybrid approach; the example shown here uses a second-order polynomial, though results are similar across different orders. Metric DepthAnythingv2\cite{depthanything} is a monocular learning-based method, with its output converted to disparity. Stereo Disparity is computed using block matching between the left and right stereo images. }
    \label{fig:KITTI_Image}
\end{figure}
To further validate our system, we evaluate our depth completion method. Compared to the commercial off-the-shelf Intel D455 stereo camera, our approach demonstrates three key advantages: reduced noise, an extended effective range, and improved structural consistency.
These improvements are evident in Fig.~\ref{fig:depth_comparison}, which compares point clouds generated by the D455 stereo camera, a classical computer vision block matching algorithm, against our depth completion results. In the double story building example, the stereo reconstruction exhibits strong wavelike noise that degrades the geometry of the guard rails, stairs, and door. In contrast, our depth completion significantly reduces noise and preserves fine structural details.
Additionally the stereo system struggles with self-similar surfaces, the D455 depth maps frequently contain holes, such as around the door and window regions. These artifacts are largely eliminated in our method. Moreover, the effective range of our approach is improved. For instance, in the middle image, the stereo baseline only partially reconstructs the L-shaped building. Leveraging the scale from stereo depth, our completion algorithm interpolates the missing structure, correctly reproducing the building’s full geometry and 90° corner.
Finally, when tested in the park environment, our method generates denser and more complete reconstructions, retaining fine details such as foliage and preserving much more of the surrounding buildings than the stereo baseline. Additional qualitative results are shown in our video. Depth completion is generalizable to both indoor and outdoor scenarios as we take the camera out of the basement. We also timed the average speed of our method as $1.5~\si{ms}$ on the Orin NX showing that not only is the quality improved, but our processing can be done in real-time onboard. 

We further quantify the performance of our depth generation method using the KITTI dataset\cite{kitti_dataset}. To emphasize accuracy in the near field—critical for collision avoidance—we evaluate disparity root-mean-square error (RMSE) rather than depth RMSE, since disparity naturally weights errors at closer ranges more heavily. The error is computed as:

\begin{equation}\label{eqn:RMSE metric}
        \frac{1}{A_m}\sum_{j=0}^W\sum_{i=0}^H \left(\mathbf{M}[i,j]\Tilde{\mathbf{d}}[i,j] -  (\mathbf{M}[i,j]\mathbf{d}[i,j]\right)^2.  
\end{equation}

$\Tilde{\mathbf{d}}$ is the ground truth disparity gathered from the LiDAR depth in the KITTI dataset\cite{kitti_dataset}. $\mathbf{d}$ is the tested method's  disparity.  $\mathbf{d}$ is generated from either a single or stereo color image gathered from the cameras. $\mathbf{M}$ is a mask image representing the valid disparities in both the ground truth and generated disparity image.  $A_m$ is the number of valid pixels in in the mask image and the estimated depth image, $\mathbf{d}$. LiDAR data is only used as a ground truth reference.
Table.~\ref{tab:depth_comp_rmse} reports the root mean square error (RMSE) across multiple KITTI scenes and various polynomial orders.  Among all methods, our hybrid depth completion consistently achieves the lowest error. In particular, the second-order (quadratic) polynomial fitting yields the most accurate results overall, although the first- and third-order variants perform comparably. By contrast, the pure monocular learning method (Metric DepthAnything) performs the worst due to the sim-to-real gap. While the strucutre is similar to our depth completion method, the scale is on the whole poor as see in Fig.~\ref{fig:KITTI_Image}. Standard stereo block matching achieves reasonable performance but is limited by structural noise and missing regions, as reflected in both the qualitative and quantitative results. These missing regions makes it unsuitable for use in a mapping context.

\subsection{Trajectory Planning Comparison}
\begin{table*}[t]
    \centering\resizebox{2\columnwidth}{!}{%
    \begin{tabular}{c|c|c|c|c|c}
        \hline
       Map  &   Volume $(\si{m^3})$ &Cuboid Decomp. $(\si{s})$ & A* on Convex Set $(\si{s})$  & RRT $(\si{s})$  & Convex Polytope (MINCO) $(\si{s})$ \\
       \hline
       Town  & $12000$ & $0.020$ & $<0.001$ & $0.005$ & $0.043$ \\
       Hospital   & $16000$ & $0.101$ & $<0.001$ & $0.600$ & $0.043$ \\
       Parking   & $500$ & $0.006$ & $<0.001$ & $<0.001$ & $0.014$ \\
       Forest   & $1000$ & $0.015$ & $<0.001$ & $0.200$ & $0.025$ \\
       Perlin Noise   & $12500$ & $0.075$ & $<0.001$ & $0.076$ & $0.040$ \\
       \hline
    \end{tabular}}
    \caption{Path Planning Timing: Cuboid Decomp is our method which is the time it takes to covert the occupancy map to a graph of convex sets. We use A* on Convex Sets to select the optimal set of safe flight corridors. RRT is rapidly exploring random tree implemented using the OMPL toolbox\cite{ompl} and used in MINCO as an alternate path planning. Convex Polytope is the Safe Flight Corridor generation method over RRT used in MINCO. }
    \label{tab:path_planning_time}
\end{table*}

\begin{table*}[t]
    \centering\resizebox{2\columnwidth}{!}{%
    \begin{tabular}{c|c|c|c|c|c|c}
        \hline
       Map Size &   $30\%$  Decomp. Time $(\si{s})$ & $30\%$ A* Time $(\si{s})$   &   $50\%$  Decomp. Time $(\si{s})$ & $50\%$ A* Time  $(\si{s})$ &   $65\%$  Decomp. Time $(\si{s})$ & $65\%$ A* Time $(\si{s})$   \\
       \hline
       $25\si{m}\times25\si{m}\times5\si{m}$  & $0.075$ & $<0.001$ & $0.14$ & $<0.001$ & $0.18$ & $<0.001$\\
         $50\si{m}\times 50\si{m}\times5\si{m}$  & $0.542$ & $0.003$ & $0.689$ & $0.003$& $0.851$ & $0.004$\\
         $75\si{m}\times75\si{m}\times5\si{m}$  & $1.75$ & $0.014$ & $2.01$ & $0.017$ & $2.24$ & $0.018$ \\
       \hline
    \end{tabular}}
    \caption{Cuboid Decomposition timing on each map is randomly generated using Perlin Noise for different infill levels and different volumes.  The A* Timing is the average timing it takes to navigate one corner of the map to another.}
    \label{tab:cuboid_decomp_scaling}
\end{table*}

We benchmark our trajectory planning framework against MINCO \cite{WANG2022GCOPTER}, a widely used global planner known for producing time-optimal, collision-free trajectories with strong computational efficiency. All evaluations are performed onboard the NVIDIA Orin NX to ensure fair comparison.
We demonstrate our results in  Fig.~\ref{fig:trajectory_path_comparison} compares paths generated by both methods across five representative maps drawn from prior works \cite{WANG2022GCOPTER,RAPTOR, xu2023vision}. Our trajectories are generally comparable in length to those produced by MINCO, though occasionally slightly longer. However, our method explicitly enforces safety constraints, whereas MINCO does not guarantee safety and can return potentially unsafe trajectories. Fig.~\ref{fig:cuboid_decomp_result} illustrates the safe flight corridors generated by our cuboid decomposition method. By tightly fitting free space around obstacles, we ensure robust feasibility even in cluttered environments and improved computation time.

We show this computation time in Table~\ref{tab:path_planning_time}  which presents a timing breakdown for different planning components. Our pipeline consists of two main steps: (i) cuboid decomposition of the occupancy map into a graph of convex sets, and (ii) A* search over this graph to select the optimal sequence of safe flight corridors. Both steps are highly efficient: decomposition can be solved in parallel before the planning occurs, and A* search completes in under 1 ms. In contrast, MINCO relies on RRT-based sampling for path discovery, which dominates its runtime. Consequently, across all test maps, our framework computes global trajectories substantially faster than MINCO as seen in Fig.~\ref{fig:trajectory_path_comparison}. Fig.~\ref{fig:trajectory_path_comparison} contains the full trajectory pipeline computation time.

Finally, we evaluate the systems scalability. Table~\ref{tab:cuboid_decomp_scaling} evaluates runtime scalability under increasing map sizes generated by Perlin noise. As expected, cuboid decomposition time scales approximately linearly with environment volume. While this may limit update frequency in very large-scale maps, the subsequent A* search remains negligible compared to decomposition. 
In summary, our trajectory planning approach consistently achieves lower computation times than MINCO while preserving comparable path quality and offering guaranteed safety constraints. These properties make it well-suited for real-time autonomous flight in cluttered and dynamically evolving environments.

\subsection{Indoor Navigation}
\begin{figure*}[t!]   \centering\includegraphics[width=\linewidth]{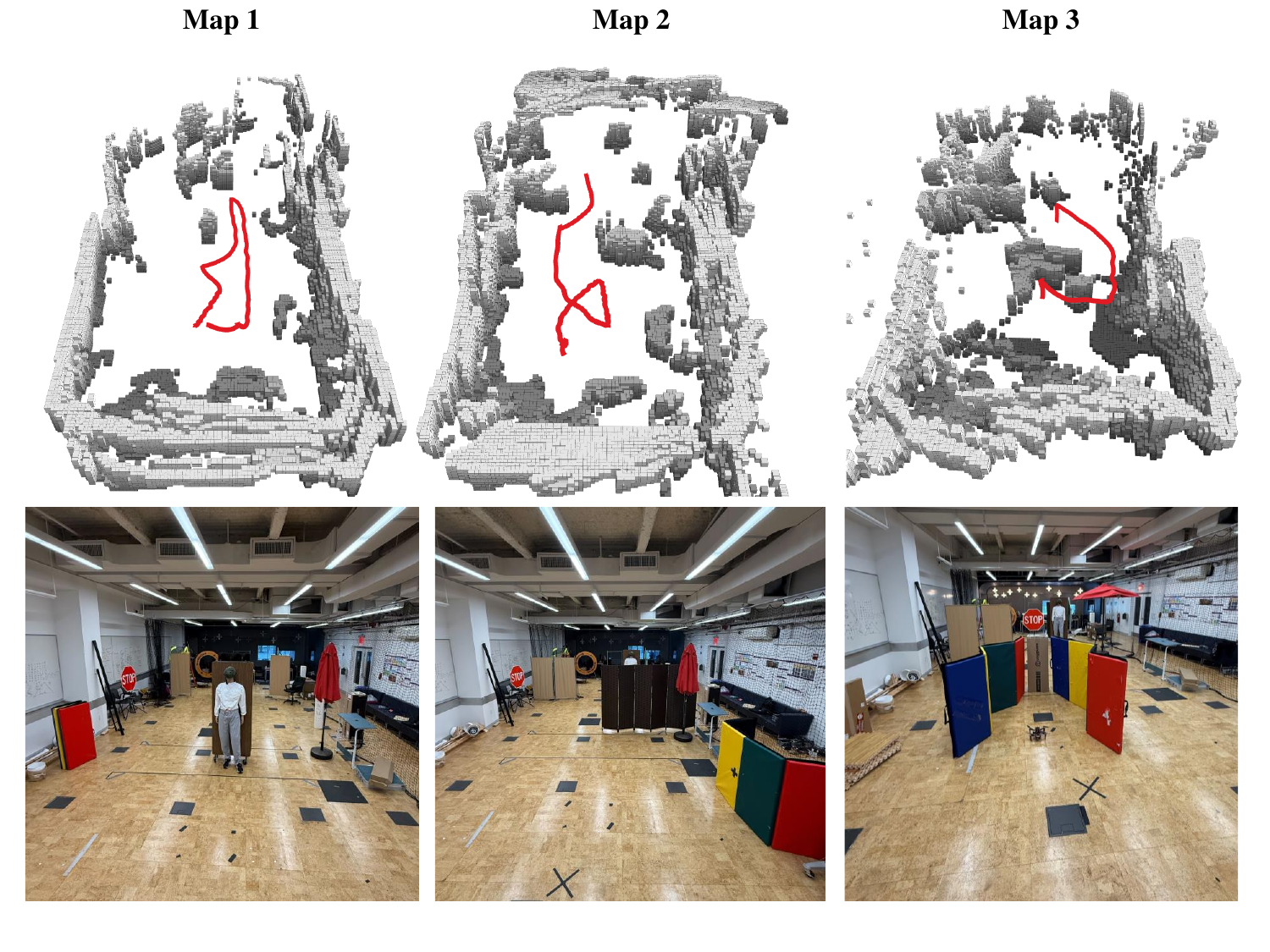}
    \vspace{-3em}
    \caption{Indoor Exploration Results: The top images include the generated occupancy map in flight. The bottom images are the color images of our map. The red path is the the path the quadrotor followed. The grayscale represents the height where darker blocks are lower and lighter blocks on top. For improved visualization the floor and ceiling are removed in our maps in post processing.}
    \label{fig:indoor_explr}
\end{figure*}

We show that our method is generalizable to indoor environments. We perform $3$ additional exploration tests presented in Fig.~\ref{fig:indoor_explr}. For map $1$ and $2$, the quadrotor explores indoors without GPS and still generates high-quality maps. 
For map $3$, we demonstrate that our quadrotor can navigate non-convex dead ends using global mapping information. Our drone is able to clear all $3$ scenarios.
In the left $2$ images, our exploration path is safe and generates a complete map.
Additionally, we see that the generated map reflects the real-world image well containing the mannequin in map $1$ and the barriers in map $2$.
In the rightmost image, the robot must navigate behind the barricade. The planner's height is constrained to prevent it from flying over the obstacle. 
Furthermore, depth completion has a weakness. Close objects can cause massive distortion in the structure of the image. This is the reason behind the lower quality of map $3$ has with respect to maps $1$ and $2$. 
Despite these challenging perception and trajectory generation problems, the quadrotor successfully escapes to the other side showing our method's robustness to sensor corruptions and tight corridors.

\section{Conclusion} \label{sec:conclusion}
In this work, we present a complete visual sensing and planning system for autonomous navigation  of unknown environments. This method is effective in both indoor and outdoor scenarios and capable of fully autonomous exploration using purely onboard sensors.
Our method performs superiorly to other popular off-the-shelf methods in terms of depth sensing, and is computationally faster while maintaining safety guarantees in terms of planning. 
However, we acknowledge that our method presents a few limitations. 
When generating our cuboid decomposition, we reprocess the full map each time to create the graph of convex sets. This can become computationally expensive as the map grows in size.
In future works, we will convert our algorithm to take incremental updates on the map to reduce the computational burden. 
Additionally, our depth completion algorithm requires an accurate metric depth to rescale. Even if our method is resilient to a noisy baseline metric depth, it is vulnerable if the baseline depth does not exist.
Future works will address these limitations and seek to improve both the depth generation and further reduce the computation time required for trajectory planning.

\section*{APPENDIX}\label{sec:appendix}
This section details the definition for the Bernstein Polynomials and explain the derivations for eq.~\ref{eqn:way_coeff_convert}~\ref{eqn:ineq_const}~\ref{eqn:bern_deriv}.
All properties are derived from the basis function, eq.~\ref{eqn:basis_func} . In this section we consider  a single spline of order $n$ and time duration $T$. 
\begin{equation}\label{eqn:basis_func}
    \begin{split}
        b_{v,n}(t) &=  \binom{n}{v} \left(\frac{t}{T}\right)^v\left(1-\frac{t}{T}\right)^{n-v} \\
        B_n(t) &= \sum_{v=0}^{n} c_{v,n} b_{v,n}(t)  ~t\in[0,T]
    \end{split} 
\end{equation}
where $v$ represents the basis or coefficient number $t$ represents time and $\binom{n}{v} = \frac{n!}{(n-v)!v!}$. Because this section needs to go over the generalized form of Bernstein Polynomial for any order, we change the definition of our coefficient subscript from above $c_{v,n}$. The following represents the $v$ coefficient of a $n$ order Bernstein polynomial. 
Using the definition of the basis function the following properties, start and end point special definitions, derivative definition, and the convex hull property will be derived.

\subsection{Bernstein Polynomial Derivative}
In this section, we will prove that the time derivative represented by the symbol $'$ for our Bernstein polynomial of order $n$ can be represented by a lower order $n-1$ Bernstein  polynomial with the following coefficients

\begin{equation}
\label{eqn:bern_coeff_deriv}
B'_n(t) = \frac{n}{T}\sum_{v=0}^{n-1}b_{v,n-1}(t)\left(c_{v+1,n}  -c_{v,n} \right).
\end{equation}

 We first take the derivatives of our basis function and get $3$ cases, $v=0,n$ and $v\in[1,n-1]$
\begin{equation} \label{eqn:basis cases}
     \begin{split}
       \text{$b'_{0,n}\left(t\right)$} &= -\frac{n}{T}\left(b_{0,n-1}(t)\right), \\
        \text{$b'_{v\in[1,n-1],n}\left(t\right)$}&= \frac{n}{T}\left(b_{v-1,n-1}(t)-b_{v,n-1}(t)\right), \\
        \text{$b'_{n,n}\left(t\right)$}&= \frac{n}{T}\left(b_{n-1,n-1}(t)\right). 
     \end{split}
\end{equation}

This result comes from the rewriting the basis function as
\begin{equation}
    \begin{split}
        b_{v,n}(t) &=  \binom{n}{v} \left(\frac{t}{T}\right)^v\left(1-\frac{t}{T}\right)^{n-v}, \\
        b_{v,n}(t) &=  \binom{n}{v}f(t)g(t), \\
        f(t) &=  \left(\frac{t}{T}\right)^v, \\
        g(t) &=  \left(1-\frac{t}{T}\right)^{n-v}, \\
    \end{split} 
\end{equation}

We can then write the time derivative as the follow values. 
\begin{equation}
        b'_{v,n}(t) =  \binom{n}{v}f'(t)g(t) +\binom{n}{v}f(t)g'(t)  
\end{equation}

We can derive the first term as \begin{equation}
       \binom{n}{v}f'(t)g(t) = 
     \begin{cases}
       \text{$0$} &\quad\text{if $v=0$}\\
        \text{$\frac{n}{T}b_{v-1,n-1}(t)$}&\quad\text{if $v>0$}
        \end{cases}.
\end{equation}

The second term forms
\begin{equation}
       \binom{n}{v}f(t)g'(t) = 
     \begin{cases}
       \text{$0$} &\quad\text{if $v=n$}\\
        \text{$-\frac{n}{T}b_{v,n-1}(t)$}&\quad\text{if $v<n$}
        \end{cases}.
\end{equation}

When we combine these two terms, we get the $3$ special cases in eq.~(\ref{eqn:basis cases}). 
By substituting the time derivative of the basis functions in our Bernstein polynomial we get the following summation

\begin{equation}
\begin{split}
        B'_n(t) = \frac{n}{T}(c_{n,n}b_{n-1,n-1}(t)\\
        +\frac{n}{T}\sum_{v=1}^{n-1}c_{v,n}\left(b_{v-1,n-1}(t) -b_{v,n-1}(t) \right)\\
        -\frac{n}{T}c_{0,n}b_{0,n-1}(t),
\end{split}
\end{equation}

We can realize this equation is exactly the same as the eq.~(\ref{eqn:bern_coeff_deriv}). We can represent this using matrix multiplication as follows 

\begin{equation}\label{eqn:bern_deriv_example}  \begin{split}
    B'_n(t) &= \frac{n}{T}
\begin{bmatrix}
b_{0,n-1}(t)  \\
\vdots\\
b_{n-1,n-1}(t) 
\end{bmatrix}^\top\begin{bmatrix}
 -1 & 1 & 0 &   \hdots  \\
 0 & -1 & 1 & \hdots \\
  &  &   \ddots & \ddots   \\
\end{bmatrix}\begin{bmatrix}
c_{0,n} \\
\vdots\\
c_{n,n}
\end{bmatrix}, 
\end{split}
\end{equation}

We can extrapolate this to a higher order time derivatives by repeatedly applying eq.~(\ref{eqn:bern_deriv_example}). This results form the core of eq.~(\ref{eqn:bern_deriv}).

\subsection{Bernstein Endpoints}

Let us recall, the basis function is constructed from products of $\frac{t}{T}$ and $1-\frac{t}{T}$. This means for the special cases $B_n(0)$ and $B_n(T)$. The  basis function is a result of powers of $1$ and $0$, This property enforces that the only active basis functions are the ones with an active $0^0$ power. This gives the following properties. 

\begin{equation}
      B_n(0) = c_{0,n},~~
      B_n(T) = c_{n,n}.
\end{equation}

Using eq.~(\ref{eqn:bern_deriv_example}), we substitute time values $0$ and $T$ to get the following 

\begin{equation}
\begin{split}
      B'(0) &= \frac{n}{T}(c_{1,n} - c_{0,n}),\\
          B'(T) &= \frac{n}{T}(c_{n,n} - c_{n-1,n}).
\end{split}
\end{equation}

Finally, we use the same logic to acquire the relationship between the start and end acceleration, $''$, of the polynomial as

\begin{equation}
\begin{split}
      B''(0) &= \frac{n(n-1)}{T^2}(c_{2,n}- 2c_{1,n} + c_{0,n}),\\
      B''(T) &= \frac{n(n-1)}{T^2}(c_{n,n} - 2c_{n-1,n} + c_{n-2,n}).
\end{split}
\end{equation}

We can simply stack the matrices together and get the following result for an order $5$ polynomial

\begin{equation}\label{eqn:coeff_spline}  
    \begin{bmatrix}
B(0)\\
B'(0) \\
B''(0) \\
B(T)\\
B'(T)\\
B''(T)
\end{bmatrix}=
\begin{bmatrix}
1 & 0 & 0 & 0 & 0 & 0\\
-\frac{5}{T} & \frac{5}{T} & 0 & 0 & 0 & 0\\
\frac{20}{T^2} & -\frac{40}{T^2} & \frac{20}{T^2} & 0 & 0 & 0\\
0 & 0 & 0 & 0 & 0 & 1\\
0 & 0 & 0 & 0 & -\frac{5}{T} & \frac{5}{T}\\
0 & 0 & 0 & \frac{20}{T^2} & -\frac{40}{T^2} & \frac{20}{T^2}
\end{bmatrix}
\begin{bmatrix}
c_{0,5}\\
c_{1,5}\\
c_{2,5}\\
c_{3,5}\\
c_{4,5}\\
c_{5.5}
\end{bmatrix}.
\end{equation}

We acquire eq.~(\ref{eqn:way_coeff_convert}) by inverting the above equation. This forms the relationship between our waypoints and the order $5$ Bernstein polynomial. 

\subsection{Bernstein Convex Hull Property}

Our goal is to prove the following theory where $l$ and $u$ are constant lower and upper bounds

\begin{equation}\label{eqn:convex_hull}
    \begin{bmatrix}
l\\
\vdots \\
l
\end{bmatrix} < \begin{bmatrix}
c_{0,n}\\
\vdots \\
c_{n,n}
\end{bmatrix} < \begin{bmatrix}
u\\
\vdots \\
u
\end{bmatrix} \Rightarrow l < B_n(t) < u, 
\end{equation}

Proving this property allows us to represent a continuous time constraint using only $n$ discrete constraints. Additionally, since eq.~\ref{eqn:way_coeff_convert} establishes the waypoint and coefficient relationship we can combine eq.~(\ref{eqn:way_coeff_convert}) and eq.~(\ref{eqn:convex_hull}) to obtain eq.~(\ref{eqn:poly_spline}).

To prove this property is true, we perform the following steps First, we sum our vectors on each side. 

\begin{equation}
 \begin{bmatrix}
l\\
\vdots \\
l
\end{bmatrix} \leq \begin{bmatrix}
c_{0,n}\\
\vdots \\
c_{n,n}
\end{bmatrix} \leq \begin{bmatrix}
u\\
\vdots \\
u
\end{bmatrix} \Rightarrow  \sum_{v=0}^n l \leq \sum_{v=0}^n c_{v,n} \leq \sum_{v=0}^n u,
\end{equation}

We then multiple each term of the summation individually by the basis function $b_{v,n}$.  Because each basis function is always $>0$ by its construction as we bind the polynomial between $t\in [0,T]$, this implies 
\begin{equation}\label{eqn:proof_ineq}
 l\sum_{v=0}^n b_{v,n}(t)  \leq \sum_{v=0}^n c_{v,n} b_{v,n}(t) \leq u\sum_{v=0}^n b_{v,n}(t) .
\end{equation}

We would like invoke the binomial theorem from algebra \cite{binom_theorem} and the basis function, eq.~\ref{eqn:basis_func}, to derive that the sum of our basis function is equal to $1$.

\begin{equation}
\begin{split}
     (x+y)^n &= \sum_{v=0}^n \binom{n}{v}x^vy^{n-v}, \\
  b_{v,n}(t) &=  \binom{n}{v} \left(\frac{t}{T}\right)^v\left(1-\frac{t}{T}\right)^{n-v},    \\
  \sum_{v=0}^n b_{v,n}(t)  &= \left(1-\frac{t}{T}+\frac{t}{T}\right)^n = 1.
\end{split}
\end{equation}

Using these values, we are able formulate the final part of our proof by replacing  $\sum_{v=0}^n b_{v,n} =1$ in eq.~(\ref{eqn:proof_ineq})
\begin{equation}
 l \leq \sum_{v=0}^n c_{v,n} b_{v,n}(t) \leq u
 \Rightarrow  l \leq B_n(t) \leq u.
\end{equation}

This means that if the coefficients are bound by a lower and upper value then the resulting function is also bound by the same coefficients. We can apply the same properties to bind the future derivatives. This forms the proof for eq.~\ref{eqn:ineq_const} and allows us to set continuous time constraints using a finite number of discrete constraints. We can also apply this to higher order derivative constraints by combining this property with the derivative property.

\subsection{Distance Metrics for Convex Sets}
\begin{figure}
    \centering
    \includegraphics[width=\linewidth]{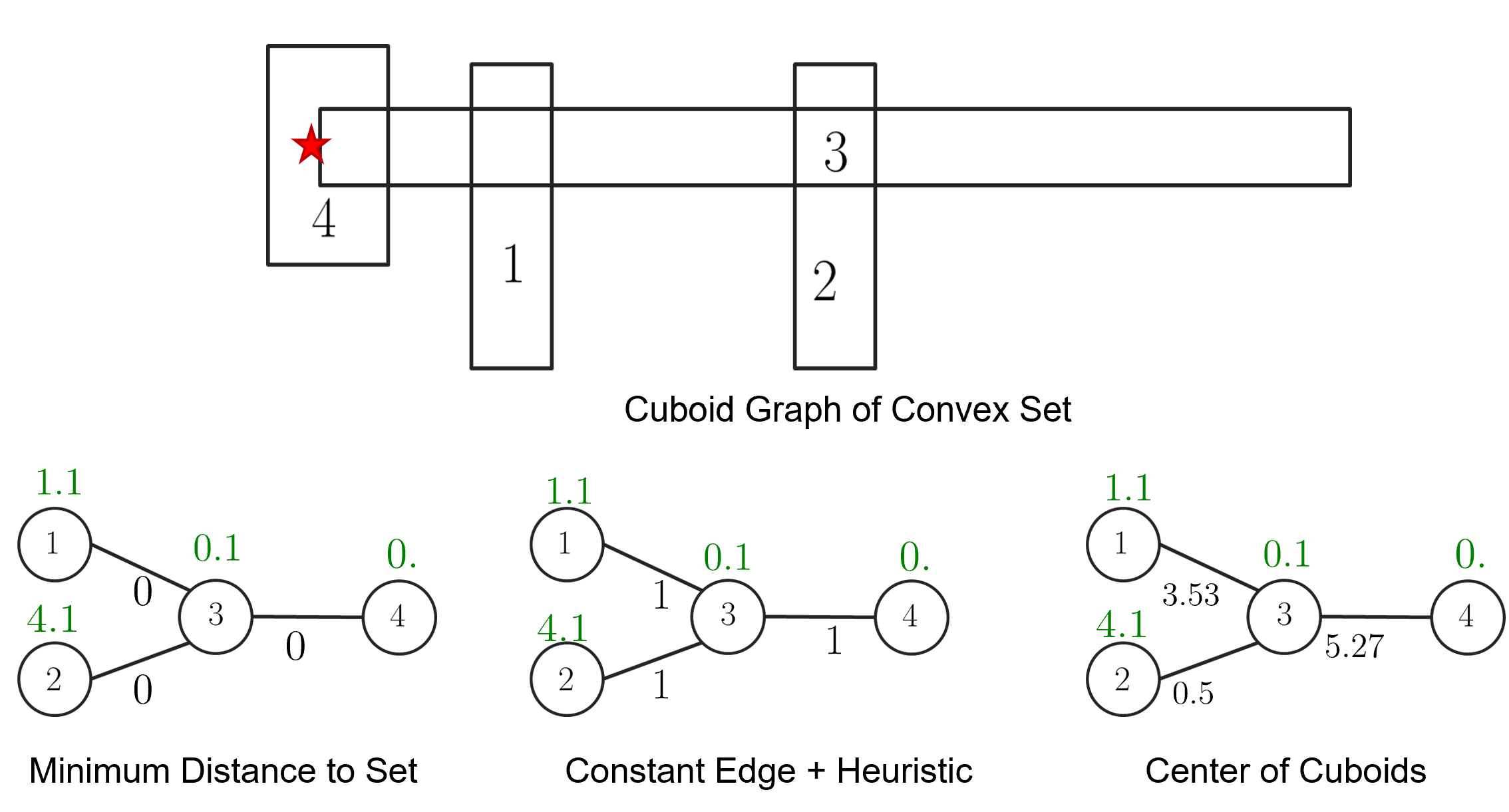}
    \caption{ The top map is a drawing of how the convex sets intersect. The bottom graphs are their abstracted from with edge weights. The green value represents the heuristic. The Star represents the target.}
    \label{fig:corridor_example}
\end{figure}

 We will motivate why distance metric between sets is difficult to define in Fig.~\ref{fig:corridor_example}. Formally, Fig.~\ref{fig:corridor_example} is defined using the following set of $2-\text{D}$ convex corridors. Each corridor is in the form $[x_{\text{min}},x_{\text{max}}]\times[y_{\text{min}},y_{\text{max}}]$
 \begin{itemize}
\item Corridor 1: $[-4,-3] \times[-2,2]$   Center $(-3.5,0)$
\item Corridor 2: $[-1,1] \times[-2,2]$  Center $(0,0)$
\item Corridor 3: $[-5,5] \times[0,1]$  Center $(0,0.5)$
\item Corridor 4: $[-6,-4.5] \times[0,2]$ Center $(-5.25,1)$
\item Target: $(-5.1,0.5)$. \end{itemize}

 We will demonstrate analyze following metrics below 

\textbf{Shortest Distance between two sets}.
This metric is useful for a heuristic for long term planning, but we see the definition of two convex sets being connected this value is always zero. We see that calculating the shortest distance between a set and the target gives good guidance for each set. This makes it unsuitable for a search metric additionally there becomes no penalty for including an arbitrary number of convex sets.

\textbf{Constant Edge Weight}.
This minimizes the number of splines or optimization variables as we naturally are minimizing the number of safe flight corridors. This does not minimize the minimum distance as we see the paths $1-3-4$ is equal to path $2-3-4$. We add the heuristic metric previously to encode some amount of optimality between two sets. The path $1-3-4$ becomes a shorter path. We found in our experimental results this is a good approximation for the maps we face. 

\textbf{Center of Cuboids}.
This metric does not have a useful physical meaning in distance. Corridors closer to the center are considered lower distance than corridors farther from the center. The distance from corridor 3 to 4 is further than the distance from the corridor 1 to 3. As a result, we see path $1-3-4$ has a much longer distance than path $2-3-4$. This is the opposite of reality. Adding the heuristic to the search still has the incorrect path $2-3-4$ be further.
\bibliographystyle{IEEEtran}
\bibliography{references}
\end{document}